\documentclass[a4paper,fleqn]{cas-dc}

\usepackage[numbers]{natbib}
\usepackage{amsthm}
\usepackage{graphicx}

\usepackage{times}
\usepackage{soul}
\usepackage{url}
\usepackage{xcolor}
\usepackage{svg}
\usepackage{amsmath}
\usepackage{cleveref}
\crefformat{section}{\S#2#1#3} % see manual of cleveref, section 8.2.1
\crefformat{subsection}{\S#2#1#3}
\crefformat{subsubsection}{\S#2#1#3}
\usepackage[utf8]{inputenc}
\usepackage[small]{caption}
\usepackage{svg}
\usepackage[final]{pdfpages}
\usepackage{graphicx}
\usepackage{amsthm}
\usepackage{booktabs}
\usepackage{algorithm}
\usepackage{algorithmic}
\usepackage{booktabs}
\usepackage[colorinlistoftodos]{todonotes}
\usepackage[inline]{enumitem}
\usepackage{subcaption}
\usepackage{tikz}
\usepackage{pgfplots}
\pgfplotsset{compat=1.18}
\usepackage{tcolorbox}

\newcount\Comments  % 0 suppresses notes to selves in text
\newcommand{\kibitz}[2]{\ifnum\Comments=1\textcolor{#1}{#2}\fi}

\theoremstyle{definition}

\def\tsc#1{\csdef{#1}{\textsc{\lowercase{#1}}\xspace}}
\tsc{WGM}
\tsc{QE}
\tsc{EP}
\tsc{PMS}
\tsc{BEC}
\tsc{DE}

\newcommand{\benchmark}{\textsc{OPTICS}}
\newcommand{\first}{\textsc{OPTICS-CNT}}
\newcommand{\second}{\textsc{OPTICS-CMP}}

\definecolor{cb}{HTML}{00B2FF}
\definecolor{cg}{HTML}{00BF63}
\definecolor{cr}{HTML}{FF3131}

\definecolor{opblue}{HTML}{004aad}
\definecolor{rcred}{HTML}{CC0000}
\definecolor{dogreen}{HTML}{2E6417}
\definecolor{typurple}{HTML}{5E17EB}

\newtcbox{\lefttag}{
  on line,
  colback=RoyalBlue!10,
  colframe=RoyalBlue!40!black,
  boxrule=0pt,
  arc=2pt,
  top=1pt,bottom=1pt,left=2pt,right=2pt,
  fontupper=\scriptsize\ttfamily
}

\newtcbox{\righttag}{
  on line,
  colback=orange!10,
  colframe=orange!40!black,
  boxrule=0pt,
  arc=2pt,
  top=1pt,bottom=1pt,left=2pt,right=2pt,
  fontupper=\scriptsize\ttfamily
}

\makeatletter
\DeclareRobustCommand\onedot{\futurelet\@let@token\@onedot}
\def\@onedot{\ifx\@let@token.\else.\null\fi\xspace}

\def\etal{\emph{et al}\onedot}

\begin{document}
\let\WriteBookmarks\relax
\def\floatpagepagefraction{1}
\def\textpagefraction{.001}
\shorttitle{A Study of Commonsense Reasoning over Visual Object Properties}
\shortauthors{Kolari et~al.}

\title [mode = title]{A Study of Commonsense Reasoning over Visual Object Properties}
% \tnotemark[1,2]

% \tnotetext[1]{This document is the results of the research
%   project funded by the National Science Foundation.}

% \tnotetext[2]{The second title footnote which is a longer text matter
%   to fill through the whole text width and overflow into
%   another line in the footnotes area of the first page.}

\author[1]{Abhishek Kolari} % Hassan Zadeh}
% \cormark[1]
\ead{abkolari08@gmail.com}

\author[1]{Mohammadhossein Khojasteh}
% \cormark[1,2]
\ead{m.khojasteh@vu.nl}

\author[2]{Yifan Jiang}
% \cormark[1,2]
\ead{yifjia@isi.edu}

\author[1]{Floris den Hengst}
\ead{f.den.hengst@vu.nl}

\author[1]{Filip Ilievski}
\cormark[1]
% \cormark[1,2]
\ead{f.ilievski@vu.nl}

\address[1]{Department of Computer Science, Vrije Universiteit, Amsterdam, The Netherlands}
\address[2]{Information Sciences Institute, University of Southern California, Marina del Rey, CA, USA}

% \author{
%   Abhishek Kolari$^\dagger$
%   \
%   Mohammadhossein Khojasteh$^\dagger$
%   Yifan Jiang$^\diamond$
%   \
%   Floris den Hengst$^\dagger$
%   \
%   Filip Ilievski$^\dagger$
%   \ \\
%   $^\dagger$Department of Computer Science, Vrije Universiteit Amsterdam, Netherlands
%   \ \\
%   $^\diamond$Information Sciences Institute, University of Southern California, Marina del Rey, CA, USA
%   \\
%   \texttt{abkolari08@gmail.com}\\
%   \texttt{\{m.khojasteh,f.den.hengst,f.ilievski\}@vu.nl, yifjia@isi.edu}
% }

\cortext[cor1]{Corresponding author}

\begin{abstract}
Inspired by human categorization, visual reasoning about object properties, such as physical attributes and functions, involves identifying and recognizing low-level details and higher-level abstractions. While current visual question answering~(VQA) studies consider multiple object properties, such as size, they typically blend perception and reasoning and lack representativeness with respect to reasoning levels and image categories, making it unclear whether and how vision-language models~(VLMs) recognize and reason about depicted objects. To this end, we introduce a systematic evaluation framework comprising images of three representative types, three reasoning levels of increasing complexity, and four object property dimensions, informed by prior work on commonsense knowledge representation and reasoning. We develop a procedure to instantiate this framework in two VQA object-reasoning benchmarks: \first, comprising 360 images paired with 1,080 multi-level, count-based questions, and \second, comprising 2.1k comparison questions. Experiments with 12 state-of-the-art VLMs in zero-shot settings reveal significant limitations relative to humans, with the best-performing model achieving below 40\% counting and 70\% comparison accuracy. While newer reasoning models perform better, a 20\% gap to human performance remains. VLMs struggle particularly with photographic images, counterfactual reasoning, physical and functional properties, and higher counts. We make the \benchmark\ benchmark data and code available to support future scalable benchmarking methods, generalized annotation guidelines, and advanced reasoning VLMs.

\end{abstract}

% \begin{graphicalabstract}
% \includegraphics{figs/grabs.pdf}
% \end{graphicalabstract}

% \begin{highlights}
% \item Research highlights item 1
% \item Research highlights item 2
% \item Research highlights item 3
% \end{highlights}


\maketitle

% ---------------------------------------------------------------
% Paper content
\section{Introduction}% (1p)}

\begin{figure}[!t]
    \centering 
    \includegraphics[width=\linewidth]{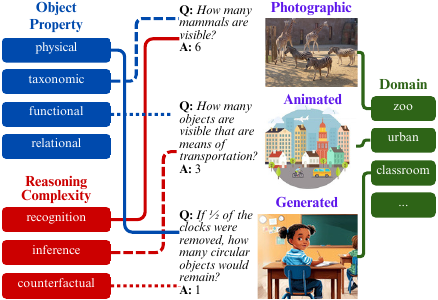}
    \caption{VQA in \textbf{\first}, with questions about different \textcolor{opblue}{object properties} of varying \textcolor{rcred}{reasoning complexity} on three \textcolor{typurple}{types of images} from different \textcolor{dogreen}{domains}.}
    \label{fig:benchmark_example}
\end{figure}

% \textbf{Background.} 
One of the core systems of human cognition is object property reasoning
%Object property reasoning is one of the core systems of human cognition
~\cite{spelke2007core}. 
% It involves identifying object properties and recognizing both low-level details and higher-level abstractions.
It involves 
% Importantly, this process of object perception involves 
a multi-level reasoning from low-level recognition, through categorization, to identification and inference
% involves the processes of 
% segmentation, categorization, and identification 
\cite{fields2016humans}.
By abstracting object properties and interactions, humans represent their environment and perform inference, a capability already present when real-world knowledge and experience are limited during infancy~\cite{lipton2004discrimination}. 
For instance, to count how many mammals are visible in \cref{fig:benchmark_example}-top, humans recognize the image objects, decide which of them fit the constraint %(type)
`type of mammal', and count those. % that satisfy this criterion.
The spatio-temporal principles of cohesion (objects move as connected, bounded wholes), continuity (objects move along connected, unobstructed paths), and contact (objects do not interact at a distance) drive such
object-property reasoning in humans. 
Humans are known to do this by perceiving object boundaries, objects
%shapes that move 
(partially) out of view, and by predicting object movement~\cite{aguiar19992,leslie1987six}.

% Conversely, understanding object properties including continuity, spatio-temporal cohesion, and causality of interactions with other objects is essential for developing intelligent computational systems with commmon sense~\cite{}.

Recent work has used visual question answering~(VQA) as a diagnostic task for vision-language models~(VLMs).% where
Here, most existing benchmarks rely on extracting and reasoning over physical and abstract object properties~\cite{johnson2017clevr,hudson2019gqa,acharya2018tallyqaansweringcomplexcounting}.
Some of %existing VQA
these benchmarks cover reasoning levels of increasing complexity, from recognition of basic visual elements such as physical attributes and taxonomic categories~\cite{yuan2021perception,tong2024eyes}, to inference of relational properties~\cite{lu2022learn,hudson2019gqa,cai2024vip}, to counterfactual reasoning~\cite{frohberg2021crass,yu2023ifqa,li2024eyes}. However, current benchmarks still exhibit two key limitations.
The first is the \textit{lack of a delineation between subtasks for object perception and reasoning},  
i.e., object detection, inference, and counterfactual reasoning are evaluated simultaneously.
While some studies have identified small object sizes, peripheral locations, and co-occurrence of distractors as causes of failure in VLMs \cite{zhang2025mllms,zhang2024exploring,alghisi2025re}, a lacking separation of subtasks hinders comprehensive diagnosis of their strengths and weaknesses.
Second, we note a \textit{limited representativeness of the evaluation dimensions covered}, with a focus on basic features such as shape and color~\cite{goyal2017making,antol2015vqa}.
% Yifan's \cite{jiang2024marvel} recent systematic evaluation of abstract visual reasoning covers different types, configurations, and shapes.
The recent systematic evaluation of abstract visual reasoning by Jiang \etal \cite{jiang2024marvel} covers more complex features, such as types, configurations, and shapes. However, it is limited to manipulating simple shapes in synthetic environments with minimal noise, leaving the ability of VLMs to perceive and reason about objects in images with varying levels of realism and noise remains underexplored.

% Overall, we observe 

% \textbf{Approach and contributions.}
We address this gap in the understanding of the capabilities of current models with \textit{a comprehensive, fine-grained evaluation of the ability of VLMs to perceive and reason over %given
a representative range of object properties, reasoning complexities, and image types}.
%We focus on number sense questions involving exact counting and approximate comparisons.
We make three contributions (\cref{fig:benchmark_example}):
\begin{itemize}
    \item[1.] \textbf{A systematic evaluation framework} for object property reasoning, consisting of three image types with varying noise and realism, three levels of reasoning complexity, and four dimensions of object properties. 

    % \item[2.] \textbf{A semi-automatic procedure} for dataset creation, we introduce \benchmark (\underline{O}bject \underline{P}roperty Reasoning \underline{T}asks for \underline{I}mage-based \underline{C}ommon\underline{S}ense Evaluation) based on our framework. Within OPTICS, we construct two VQA benchmarks: \textbf{\first} with 1,080 high-quality counting questions about 360 images, and \textbf{\second} with 2.1k comparison questions about image pairs. %that are systematically sampled or generated.
    \item[2.] \textbf{A semi-automatic procedure} for dataset creation, resulting in 
    % introducing \benchmark\ (\underline{O}bject \underline{P}roperty Reasoning \underline{T}asks for Evaluating \underline{I}mage-based \underline{C}ommon\underline{S}ense) split into
    two VQA benchmarks based on our framework: \textbf{\first} with 1.1k high-quality counting questions about 360 images, and \textbf{\second} with 2.1k comparison questions about image pairs.
    
    %\footnote{\benchmark\ stands for \underline{O}bject \underline{P}roperty Reasoning \underline{T}asks for Evaluating \underline{I}mage-based \underline{C}ommon \underline{S}ense.} %that are systematically sampled or generated.

    \item[3.] \textbf{A fine-grained experimental analysis} of frontier VLMs on \benchmark% with in a zero-shot setting
    , which pinpoints %fine-grained
    limitations of current VLMs for photographic images, counterfactual reasoning, functional knowledge, and higher counts.
    
    % which revealed the models' performance to be only moderately higher than the random chance baseline, with further insights from dimension and structure-level analysis, and dimension-aware tuning. %We also evaluate a spatially-aware fine-tuned method \cite{Chen_2024_CVPR}, which highlights the promise of specialised reasoning models for object property tasks. 
    
\end{itemize}
% the object property reasoning problem contains three levels: the first question (direct recognition) involves locating and identifying turtles, followed by semantically classifying them as reptiles (taxonomic). The second question (property inference) demands abstraction and reasoning about the utility of a log (functional), and the third question (counterfactual) requires making hypothetical inferences about reptiles (taxonomic) based on changes in the scene (more examples included in Appendix \cref{sec: benchmark_examples}). The addition of numerical reasoning to such questions makes these components foundational to higher-order visual reasoning. To this end, our contributions are as follows:

% 1. EXPAND INTRODUCTION\\
% 2. CHECK OTHER BENCHMARK PAPERS ON DISCUSSION IDEAS AND IF IT IS THE SAME AS THE CONTENT MENTIONED IN THE RELATED WORK\\
% 3. ADD CONTRIBUTIONS TO THIS PAPER\\
% 4. CHECK IF RESEARCH QUESTIONS ARE MENTIONED EXPLICITLY OR PART OF THE INTRODUCTION

We believe these contributions will aid practitioners in applying VLM systems and researchers in focusing their future efforts. We make our code and data available at \url{https://github.com/AbhishekKolari/OPTICS}.
\begin{table*}[!t] % Start the floating table environment
\caption{Comparison of \textbf{\benchmark} to closely related existing benchmarks: VQAv2 \cite{goyal2017making}, CLEVR \cite{johnson2017clevr}, TallyQA \cite{acharya2018tallyqaansweringcomplexcounting}, SCIENCEQA \cite{lu2022learn}, GQA \cite{hudson2019gqa}, C-VQA \cite{zhang2024if}, CFMM \cite{li2024eyes}, BLINK~\cite{fu2024blink}, and CAPTURe~\cite{pothiraj2025capture}.} % Caption should be above the table
\label{tab:benchmark_comparison} % Add a label for cross-referencing
\centering % This command centers the content within the table environment
% Adjust column separation. Default is usually 1pt. Try a larger value.
\setlength{\tabcolsep}{2pt}
\small
% Use \makebox to create a box that spans the text width (\linewidth or \textwidth)
% and centers its content. If the content is wider, it will overflow symmetrically.
% \makebox[\linewidth]{
% Shrink the table to exactly the text width
% \resizebox{\textwidth}{!}{%
    \begin{tabular}{ll|c|c|c|c|c|c|c|c|c|c}
    \toprule
    \multicolumn{2}{c|}{Category} & VQAv2 & CLEVR & TallyQA & SCIENCEQA & GQA & C-VQA & CFMM & BLINK & CAPTURe &\textbf{\benchmark} \\
    \midrule
    \multirow{2}{*}{\textbf{Question}} & Counting & \textcolor{green}{\checkmark} & \textcolor{green}{\checkmark} & \textcolor{green}{\checkmark} &  & \textcolor{green}{\checkmark} & \textcolor{green}{\checkmark} & \textcolor{green}{\checkmark} & \textcolor{green}{\checkmark} & \textcolor{green}{\checkmark} & \textcolor{green}{\checkmark} \\
    & Comparison &   & \textcolor{green}{\checkmark} & & \textcolor{green}{\checkmark} & \textcolor{green}{\checkmark} & & \textcolor{green}{\checkmark} & \textcolor{green}{\checkmark} & &\textcolor{green}{\checkmark}\\
    \midrule
    % \multirow{4}{*}{\textbf{Object Properties}}
    & Physical & \textcolor{green}{\checkmark} & \textcolor{green}{\checkmark} & \textcolor{green}{\checkmark} & \textcolor{green}{\checkmark} & \textcolor{green}{\checkmark} & \textcolor{green}{\checkmark} & \textcolor{green}{\checkmark} & \textcolor{green}{\checkmark} & & \textcolor{green}{\checkmark} \\
     \textbf{Object} & Taxonomic &  &  &  & \textcolor{green}{\checkmark} & \textcolor{green}{\checkmark} &  &  & \textcolor{green}{\checkmark} & & \textcolor{green}{\checkmark} \\ 
     \textbf{Property} & Functional &  &  &  & \textcolor{green}{\checkmark} &  &  &  & \textcolor{green}{\checkmark} & & \textcolor{green}{\checkmark} \\ 
     & Relational &  & \textcolor{green}{\checkmark} & \textcolor{green}{\checkmark} &  & \textcolor{green}{\checkmark} &  & \textcolor{green}{\checkmark} & \textcolor{green}{\checkmark} & \textcolor{green}{\checkmark} & \textcolor{green}{\checkmark} \\
    \midrule
    \multirow{3}{*}{\textbf{Image Type}} & Real & \textcolor{green}{\checkmark} & \textcolor{green}{\checkmark} & \textcolor{green}{\checkmark} & \textcolor{green}{\checkmark} & \textcolor{green}{\checkmark} & \textcolor{green}{\checkmark} & \textcolor{green}{\checkmark} & \textcolor{green}{\checkmark} & \textcolor{green}{\checkmark} & \textcolor{green}{\checkmark} \\
     & Animated &  &  &  & \textcolor{green}{\checkmark} &  &  &  & & & \textcolor{green}{\checkmark} \\
     & AI-Generated &  &  &  &  &  &  & &  & \textcolor{green}{\checkmark} & \textcolor{green}{\checkmark} \\
    \midrule
    \multirow{3}{*}{\textbf{Reasoning}} & Direct Recognition & \textcolor{green}{\checkmark} & \textcolor{green}{\checkmark} & \textcolor{green}{\checkmark} & \textcolor{green}{\checkmark} & \textcolor{green}{\checkmark} & \textcolor{green}{\checkmark} & \textcolor{green}{\checkmark} & \textcolor{green}{\checkmark} & & \textcolor{green}{\checkmark} \\
     & Property Inference &  &  &  & \textcolor{green}{\checkmark} & \textcolor{green}{\checkmark} &  & & \textcolor{green}{\checkmark} & \textcolor{green}{\checkmark} & \textcolor{green}{\checkmark} \\
     & Counterfactual &  &  &  &  &  & \textcolor{green}{\checkmark} & \textcolor{green}{\checkmark} & & \textcolor{green}{\checkmark} & \textcolor{green}{\checkmark} \\
    \bottomrule
    \end{tabular}
% } %
% } % End of \makebox
\end{table*}

\section{Related Work}% (0.75p)}
\noindent \textbf{Object Properties in VQA.}  
VQA has long served as a diagnostic task for evaluating the perceptual and reasoning capabilities of AI systems. Early efforts~\cite{antol2015vqa,goyal2017making} emphasized general visual recognition of objects, colors, and basic attributes, while subsequent benchmarks~\cite{singh2019towards,mathew2021docvqa,lu2022learn} targeted domain-specific context. 
Although many benchmarks across domains~\cite{johnson2017clevr,hudson2019gqa,acharya2018tallyqaansweringcomplexcounting} require reasoning over object properties, they seldom provide a systematic separation of object property dimensions~\cite{fu2024blink}. Our framework covers four distinct object dimensions based on commonsense resources~\cite{speer2017conceptnet,ilievski2021dimensions,kurtz2021object}: physical, taxonomic, functional, and relational, and enables fine-grained assessment of object-level reasoning across image types and reasoning categories, as visualized in \cref{tab:benchmark_comparison}. Moreover, while both counting and comparison questions have been relatively well-covered in prior work, our evaluation enables compositional evaluation by comparing counts rather than object properties %(e.g., which object is larger
~\cite{li2024eyes}
%)
, which better aligns with studies of number sense in humans and LLMs~\cite{dehaene2011number,thawani2021representing}.

\noindent \textbf{Visual Reasoning Complexity.}
Human cognition is inherently compositional and integrates multiple scene aspects into higher-level reasoning, with complexity progressing from direct recognition to property-level inference, and counterfactual reasoning~\cite{hoffman1987parts,xu2021sutd}. Initial \textit{direct recognition} is limited to recognition of basic visual elements such as physical attributes or taxonomic category membership~\cite{yuan2021perception,tong2024eyes}. \textit{Inference} builds on recognition, targets higher-level functional or relational properties, and requires multi-step abstraction beyond surface-level features~\cite{lu2022learn,hudson2019gqa}. \textit{Counterfactual reasoning} involves reasoning about hypothetical changes made in inputs. This type of reasoning is most challenging, as it requires understanding and adapting to hypothetical, altered, or out-of-context scenarios~\cite{frohberg2021crass,yu2023ifqa,li2024eyes,pothiraj2025capture}. Benchmarks that target different types of reasoning exist (\cref{tab:benchmark_comparison}), yet none systematically test all three reasoning levels simultaneously. 
Inspired by compositional evaluation frameworks in abstract visual reasoning~\cite{jiang2024marvel}, \benchmark\ is the first to encompass all three reasoning types and go beyond previous work to study VLM performance across
% extend evaluation to
%three
image types, thereby enabling a more comprehensive assessment of models’ perception and reasoning abilities. To support fair evaluation, we focus on counting and comparison questions in line with prior benchmarks~\cite{acharya2018tallyqaansweringcomplexcounting,johnson2017clevr,li2024eyes}. 

\section{A Commonsense Framework for Evaluating Object Property Reasoning}% (1.25)}
\label{sec:framework}

We propose a systematic framework~(\cref{fig:benchmark_example}) for the rigorous evaluation of VLMs, integrating \textit{object property dimensions}, \textit{levels of reasoning complexity}, and diverse \textit{image types}. Each dimension and its underlying design rationale are detailed in the following subsections.
% This section 
% We next detail its three components. % of the framework.
% As a multidimensional benchmark for VQA, \benchmark covers different object property dimensions (Section \cref{ssec: prop_dimensions}), a variety of image types (Section \cref{ssec: image_types}), and different levels of reasoning complexity (Section \cref{ssec: reasoning_levels}). The benchmark composition is shown in Figure \cref{fig:benchmark_compostion}. The data curation process is presented in Section \cref{sec: data_curation}. Category-specific examples from the benchmark are included in Appendix \cref{sec: cat_examples}.

\subsection{Object Property Dimensions}
\label{ssec: prop_dimensions}

Drawing on research \cite{ilievski2021dimensions,tandon-etal-2017-webchild} on representing commonsense knowledge about objects, our framework identifies four dimensions essential for counting and comparison in VQA. These dimensions reflect how humans conceptualize and reason about object attributes and relationships within visual scenes. 
%demanding that models not only identify instances, but also understand nuanced object-level concepts and their relations, thereby enabling the assessment of 
% These dimensions are crucial for count-based VQA, as they 
% assess the generalization and abstraction abilities in vision-language models.
%We now introduce the object property dimensions in \benchmark.%(examples in Appendix \cref{ssec: objprop_examples}).
% The \benchmark benchmark primarily focuses on evaluating model performance across four major object property dimensions, each of which captures a distinct facet of human conceptualisation and reasoning about objects in visual scenes. In the context of count-based visual question answering (VQA), these object property dimensions hold relevance. Models are required to identify instances and also understand nuanced object-level concepts to produce correct answers, thereby enabling assessment of generalisation and abstraction abilities in vision-language models. Example questions are included in Appendix \cref{ssec: objprop_examples}. The object property dimensions are organised into the following labels:

\noindent \textbf{Physical} knowledge 
% This dimension 
refers to the 
% surface-level, perceptual features of an object 
% and includes aspects 
% that are 
dynamics of physical systems based on observable phenomena and fundamental principles~\cite{mccloskey1983intuitive}
% directly observable or measurable from visual cues in isolation . 
Physical properties, including object qualities (e.g., \textit{circular} in \cref{fig:benchmark_example}), materials, and part-whole relations, are prevalent in commonsense knowledge representation \cite{tandon-etal-2017-webchild,fleming2017material}.
Our framework extends the list of physical properties in prior VQA studies, such as shape, color, and size (\cref{tab:benchmark_comparison}), 
% Examples from existing work in  include simple properties such as the shape, color, and size of an object. \benchmark adds to these
with more complex attributes such as the \emph{material} (wood, metal), \emph{state} of the object (liquid, solid), and \emph{structural characteristics} of the object -- including part-whole relationships such as \textit{has wheels}. % and \textit{has legs}.   %, whereas the extensions help to evaluate systems on tasks associated with recognizing materials~\cite{fleming2017material}.

% \noindent \textbf{D1: Physical. } This dimension refers to the inherent, perceptual features of an object—those that are typically observable and measurable from visual cues. Prior VQA benchmarks (Table \cref{tab:benchmark_comparison}) that deal with object recognition all contain questions on the physical property of the object but are limited to more common and simple attributes such as the shape, color, and size of the object. Besides common attributes, \benchmark includes more complex attributes like the \textbf{material} (e.g., wood, metal), \textbf{state} (e.g., solid, liquid), and \textbf{structural characteristics/part-whole relations} (e.g., has wheels, has legs) of the object. The inclusion of \textit{materials} as a complex attribute caters to its diversity, mutability, and the challenges it poses to vision science (\cite{fleming2017material}). Questions targeting physical properties require models to ground their understanding in surface-level appearance and tangible qualities, such as "\textit{How many wooden objects are present?}" or "\textit{Count the number of objects that have legs}", a key prerequisite for basic perception-level VQA tasks. In cognitive science, the physical property dimension is inspired by the \textit{similarity}, \textit{part-whole}, and \textit{quality} relations commonly found in commonsense knowledge representations (\cite{ilievski2021dimensions}).

\noindent \textbf{Taxonomic } 
% The taxonomic dimension 
knowledge captures an object's semantic category or class membership, typically expressed as \textit{is-a} relations~\cite{ilievski2021dimensions}. Examples in our framework include broad ontological groupings, e.g., \emph{biological} (e.g., mammals, reptiles), \emph{artifact} (e.g., furniture, tools), and \emph{food} (e.g., fruits, vegetables) categories. Unlike physical attributes, taxonomic properties may not be visible from visual input alone: to determine how many mammals are present in an image (\cref{fig:benchmark_example}), recognition of species must be combined with taxonomic knowledge.
The taxonomic dimension aligns with \textit{taxonomic semantic systems} from cognitive science and domain-specific knowledge representation, which group entities based on shared features and categorical similarity~\cite{mirman2017taxonomic}. 

\noindent \textbf{Functional } properties express
 % capabilities and general design aspects of an object. This includes 
 attributes such as \emph{utilities} and \textit{capabilities} of an object (e.g., means of transportation in \cref{fig:benchmark_example}), % can be worn, can cut, can hold liquid), 
\emph{affordances} as actions an object receives (e.g., breakable, foldable), and \emph{operational dependencies} for an object to function (e.g., electricity, battery). This dimension %often requires commonsense reasoning and latent world knowledge, 
primarily aligns with the \textit{utility} %and \textit{temporal} 
and \textit{causal}
relations identified in commonsense knowledge frameworks~\cite{speer2017conceptnet,heindorf2020causenet,ilievski2021dimensions}. Questions about the utility of objects have been included in VQA studies (\cref{tab:benchmark_comparison}): for example, the OK-VQA benchmark \cite{marino2019ok} tests for knowledge about object 
% implicit functionality based on 
utility, e.g., 
% recognizing that 
\textit{a broom} is for \textit{sweeping}.

\noindent \textbf{Relational}
% This dimension
knowledge captures how objects interact, how they can be contextually grouped, and how they are situated relative to each other within a visual context. This dimension includes \textit{spatial relationships} (e.g. "\textit{How many objects are \underline{hanging from} the wall?}") and \textit{contextual grouping relations} (as in "\textit{How many \underline{couples} are visible?}") \cite{ilievski2021dimensions}.
% Questions like "\textit{How many couples are visible?}" or "\textit{How many objects can be seen hanging from the wall?}" evaluate a model's ability to reason about such properties. 
Its emphasis on relations between multiple object instances complements 
% The relational dimension is motivated by the \textit{spatial} relations and \textit{contextual} grouping relations of multiple object instances in commonsense knowledge research \cite{ilievski2021dimensions}, thus complementing 
% Its focus on spatial relations and the contextual grouping of multiple object instances complements 
the focus on properties of single objects in taxonomic and part-whole physical properties.

\subsection{Reasoning Complexity}
\label{ssec: reasoning_levels}
The four object property dimensions can support questions with varying reasoning complexity.
% of interest, we now turn to the level of complexity of questions we pose about questions. 
We introduce three levels of reasoning complexity: \emph{direct recognition}, \emph{property inference}, and \emph{counterfactual reasoning}, which are exemplified in \cref{fig:benchmark_example}. These levels are grounded in cognitive science %work
on compositional reasoning \cite{fields2016humans}.
We hypothesize that the progression from direct recognition to counterfactual reasoning requires an increasingly high level of abstraction to answer a question, and a shifting emphasis from perception to reasoning, thus increasing in difficulty~\cite{fields2016humans,xu2021sutd}.
% The level of difficulty in our framework is structured by progressively increasing levels of reasoning based on the extent of abstraction required to answer the question. % and further detailed in this section.

% To incorporate the various object property dimensions, a general question structure with progressive difficulty levels is defined. We decide the reasoning levels based on the assumed extent of abstraction required, which varies depending on the property dimension and visual understanding. From Figure \cref{fig:benchmark_example}, the assumed order of complexity is direct recognition < property inference < counterfactual. Example questions are included in Appendix \cref{ssec: reasoning_complexity_examples}. The levels are organised into the following labels:

\noindent \textbf{Direct Recognition} 
% Direct Recognition 
questions ask about features that can be detected by observation and basic taxonomic knowledge.
These questions can be answered with an atomic step of perception, and reside at the physical and taxonomic property dimensions. 
% Since answers can be derived from specific visual regions or patterns, recognition questions primarily assess object recognition abilities, 
For example, the top question in \cref{fig:benchmark_example} only requires identifying the land animals as mammals.% as mammals.

% \noindent \textbf{R1: Direct Recognition. } 
% This question structure evaluates a model’s ability to recognise or retrieve directly observable features from the image, involving minimal reasoning effort. Questions at this level are focused on the property dimensions \textbf{D1} (physical) and \textbf{D2} (taxonomic), which can be identified through a single step of perception. Since the answers can be derived from specific visual regions or patterns, this level serves as a useful way to assess the foundational recognition abilities of VLMs.

\noindent \textbf{Property Inference} 
% Property inference 
questions target the functional and relational property dimensions, % D3 (functional) and D4 (relational), 
which require deeper abstraction, introducing multi-step reasoning. This examines abilities to generalize beyond surface-level features by focusing on interdependencies and relationships among objects. For example, in \cref{fig:benchmark_example}, the middle question requires identifying a bike, a car, and a plane, categorizing them as vehicles, and then reasoning about their utility in transportation.

% \noindent \textbf{R2: Property Inference. } 
% This level contains questions targeted at property dimensions \textbf{D3} (functional) and \textbf{D4} (relational), which require deeper abstraction, introducing multi-step reasoning. Due to the properties having codependencies among objects and underlying relations across different image regions, this would push models to generalise beyond surface-level features.\\

\noindent \textbf{Counterfactual Reasoning}
% This question structure 
questions probe the ability to reason about hypothetical, altered, or out-of-context scenarios. In addition to visual recognition and inference, counterfactual questions also require commonsense knowledge and flexible abstraction (e.g., creating a contextualized procedure that identifies remaining circular objects besides half of the clocks in the bottom question of \cref{fig:benchmark_example}), thus forming the most complex reasoning level in our framework. Counterfactual questions in OPTICS span any of the four property dimensions.
% more challenging than 
% % contributes to their challenging nature and their placement above 
% property inference.

% \noindent \textbf{IMAGE TYPES}\
\subsection{Image Types}
\label{ssec: image_types}
To study generalization across visual domains, we cover three image types, expecting that accuracy varies %(often worsens)
when moving from photographic to AI-generated images due to the increased level of detail and noise \cite{johnson2017clevr,hudson2019gqa}. %This section details the types of images included.
% Examples are provided in Appendix \cref{ssec: image_type_examples}.

% According to insights from benchmarks such as CLEVR (synthetic) (\cite{johnson2017clevr}) and GQA (real) (\cite{hudson2019gqa}), 
% According to insights from synthetic (CLEVR; \cite{johnson2017clevr}) and real (GQA, \cite{hudson2019gqa}) benchmarks, models perform well on synthetic datasets due to minimal noise but generalise poorly to real-world images. To assess the generalisation ability of VLMs across visual domains, \benchmark includes a diverse set of image types (examples included in Appendix \cref{ssec: image_type_examples}) under the following labels:

\noindent \textbf{Photographic} images represent naturalistic settings and feature complex textures, lighting variations, and cluttered scenes. They are realistic, i.e., all depicted objects are found in the real world, and pose visual grounding challenges, including occlusions, viewpoint variation, and natural ambiguity. %, requiring an effective visual grounding. % from models. 
%The images depict a selection of scenes, such as street views, household interiors, and natural environments.
% \noindent \textbf{I1: Real. } 
% This type includes photographic images captured from the real world that reflect naturalistic settings. They contain complex texture, variation in lighting, and cluttered scenes displaying everyday human visual perception. With these images, challenges like occlusions, viewpoint variation, and natural ambiguity arise, requiring models to operate with a higher level of visual grounding. Examples of this type include street scenes, household interiors, or natural environments.

\noindent \textbf{Animated} images are simplified or stylized and contain minimal visual noise, as well as limited levels of detail when compared to photographic images. Humans typically create animated images in digital form, and their objects are not necessarily realistic.
Object boundaries in animated images may be easier to detect, but exaggerated representations may pose an additional challenge. 

% \noindent \textbf{I2: Animated. } Images that fall under this type are often simplified or stylised, typically eliminating visual noise and possibly resulting in a lack of realistic detail. This simplification could either aid or hinder reasoning since models may not have trouble detecting object boundaries yet struggle interpreting exaggerated representations. As shown in Table \cref{tab:benchmark_comparison}, the benchmark SCIENCEQA (\cite{lu2022learn}) contains diagrammatic images in their dataset but does not showcase explicit model performance comparison between natural and stylised images. Instances of this type in \benchmark include cartoon sceneries, game-like environments, or illustration art.

\noindent \textbf{AI-generated} images include both hyper-realistic and stylized synthetic images by %prompting
generative models. They test robustness to domain shifts by deviating from visual conventions such as real-world physics and object proportions. %, and other . 
% Due to current limitations in precise object counting with generative models, 
AI-generated images typically feature simpler compositions and include implausible objects, such as half-animals or floating glasses~\cite{kamali2024distinguish}. % without a bottom).
% \noindent \textbf{I3: AI-generated.} The recent improvements in image synthesis by generative models led to the inclusion of this image type containing both realistic and stylised illustrations. These images test a model's strength to domain shifts, as they may not follow typical real-world physics, object proportions, or visual conventions. Instances of this type follow similar themes of both type \textbf{I1} (real) and \textbf{I2} (animated) but are AI-generated.

% \begin{figure*}[t!] % [h!] is a placement specifier: here, top, bottom, page of its own
%     \centering % Centers the image horizontally on the page
%     \includegraphics{img/orbit-horizontal.pdf} % Replace with your image file name
%     \caption{Process pipeline for dataset curation, consisting of image collection, candidate QA formulation, and quality assurance steps.}% : Contains four phases Image Collection, Prompt-Guided Question Generation, Answer Annotation and Benchmark Assembly.} % Caption for the image
%     \label{fig:data_curation} % Label for cross-referencing (e.g., "as seen in Figure \cref{fig:simple_image}")
% \end{figure*}

\begin{figure}[!t] % [h!] is a placement specifier: here, top, bottom, page of its own
    \centering % Centers the image horizontally on the page
    \includegraphics[width=\linewidth]{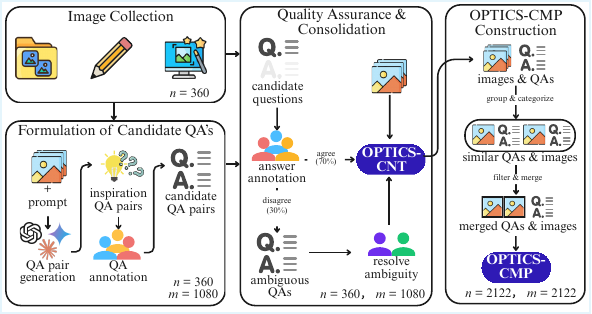} % Replace with your image file name
    \caption{\first\ and \second\ construction pipeline: image collection, candidate QA formulation, quality assurance, and pairwise combination. \textit{n} and \textit{m} are number of images and QAs, respectively.}% : Contains four phases Image Collection, Prompt-Guided Question Generation, Answer Annotation and Benchmark Assembly.} % Caption for the image
    \label{fig:data_curation} % Label for cross-referencing (e.g., "as seen in Figure \cref{fig:simple_image}")
\end{figure}

\section{Benchmark Construction}
\label{sec: data_curation}
We scope our benchmarking focus on \textit{counting} and \textit{comparison} questions to explore the three dimensions from the previous section (object properties, reasoning complexity, and image types) in depth. While these two question types are connected (comparison questions concern object counts), they capture different facets of \textit{number sense} \cite{dehaene2011number}.
Namely, comparison questions are inspired by the human ability to make quick judgments about which of two sets of objects is more numerous without establishing the exact magnitude. Thus, the focus of the comparison questions is on approximate counts.
Meanwhile, counting questions that require exact answers are non-linear in terms of human difficulty. Human perception of numeric scales is rather compressed, which makes the differences between small numbers larger than those of larger numbers (e.g., students decide that 3 is greater than 2 more quickly than deciding that 9 is greater than 8)~\cite{holyoak2025human}. OPTICS is designed to evaluate such number sense phenomena.
% in VLMs.

Our dataset curation follows a four-phase pipeline (\cref{fig:data_curation}). We \textit{collect images} from public sources corresponding to the three image types. % (I1-I3). 
We ask an annotator to \textit{formulate candidate question-answer} pairs for all images, inspired by AI-generated questions. % used as inspiration.
% prompt an MLLM to generate candidate question-answer pairs, which are manually reviewed by annotators. 
Then, the questions and answers undergo rounds of human refinement for \textit{quality assurance}, which results in \first. \second\  is then \textit{systematically constructed} from samples in \first.

\noindent \textbf{Image Collection.} To ensure diagnostic quality and mitigate data contamination risks from existing benchmarks \cite{dodge2021documenting}, we collect previously unsourced images. 
We define 26 domains based on the MacGyver dataset~\cite{tian2025macgyverlargelanguagemodels}, prioritizing domains with a greater expected variety of objects. The domains are grouped into three categories: indoors (e.g., kitchen), neutral (e.g., tools), and outdoors (e.g., zoo)
% select 15 domains, grouped into three categories, from the 
% MacGyver dataset~\cite{tian2025macgyverlargelanguagemodels}, and add 11 for a total of 26 domains 
(the complete list is given in \cref{tab:image_theme_tags}). %(e.g., "kitchen", "garage", "the zoo") 
We use these domains as queries across all three image types, enabling fair comparisons of comparable scenes.
We collect public photographic and animated images 
% \textbf{in the  domain?} 
by querying Google Images~\cite{googleimages}, Unsplash~\cite{unsplash}, and Freepik~\cite{freepik}. We use GPT-4o~\cite{openai2024gpt4ocard} and Grok3~\cite{xai2025grok3} to synthesize images (\cref{tab:aigen_prompts} shows example generative AI prompts). 
We explicitly exclude diagrammatic images~\cite{lu2022learn}. In total, we collect 360 images (120 per type).

\noindent \textbf{Formulation of Candidate QA-pairs.} 
We formulate initial question-answer pairs using a semi-automatic procedure. First, we prompt a multimodal large-language model (MLLM) to generate three counting questions (one per reasoning level) following our framework defined previously.
% in the same format structure as \benchmark
% (Figure \cref{fig:benchmark_example}),
% (see Appendix for the prompt template). 
The MLLM is selected at random from 
%three closed-source models:
% Each collected image was processed by a MLLM randomly selected at random from the 
GPT-4o \cite{openai2024gpt4ocard}, Claude 3.7 Sonnet \cite{anthropic2025}, and Gemini 2.0 Flash \cite{geminiteam2025geminifamilyhighlycapable}. 
% These MLLMs, generated questions in \benchmark's specified format using a template (Appendix \cref{ssec: Appendix_prompt_question}). 
The resulting 1,080 question-answer pairs are then refined by five human annotators (each responsible for 72 images), instructed to revise or replace the QA pairs to ensure high quality: the question should be precise, the answer accurate, and the answer count $\leq$ 10. 
% , and fits the restrictions of \benchmark (see previous section). 
% Each of the annotators is 
% A pool of five annotators annotates the images, each
% responsible for 72 images. 
The annotators are instructed to abstract over the specific object types when possible, e.g., preferring ``how many objects are made of rubber?'' over ``how many tires are made of rubber?''. 
% In practice, we used MLLM-generated questions to construct an initial structure, which annotators then refined according to each image’s details to produce the final golden dataset.
% In practice, the annotators rewrote a vast majority of the questions and nearly all answers, as many MLLM-generated questions were ambiguous or repetitive, and most answers were incorrect.

% \Cref{tab:reasoningcompelxity_questions} shows examples of the questions asked at the different reasoning levels. %Direct recognition has questions from the physical and taxonomical object property dimensions; property inference has questions from the functional and relational dimensions; and counterfactual reasoning has questions from any of the four property dimensions.

\begin{table*}[!ht]
\caption{Question examples for different reasoning complexity levels in \first: direct recognition, property inference, and counterfactual reasoning. Direct recognition has questions from the physical and taxonomical object property dimensions; property inference has questions from the functional and relational dimensions; and counterfactual reasoning has questions from any of the four property dimensions.}
\label{tab:reasoningcompelxity_questions}
\centering
\small
% Adjust column separation. Default is usually 1pt. Try a larger value.
\setlength{\tabcolsep}{5pt}
% \resizebox{\textwidth}{!}{%
% \makebox[\linewidth][c]{%
\begin{tabular}{c|c|p{0.6\linewidth}}
\toprule
\multirow{1}{*}{\textbf{Complexity}} & \textbf{Dimension} & \textbf{Question Example}\\
\midrule
\multirow{2}{*}{Direct Recognition} & physical & How many objects made of wood are present?\\ & taxonomic & How many mammals are visible in the image?\\
\midrule
\multirow{2}{*}{Property Inference} & functional & Count the number of breakable items?\\ & relational & How many objects are visible that are attached to the wall or ceiling?\\
\midrule
\multirow{4}{*}{Counterfactual R.} & physical & If one of the metal objects were replaced by a wooden object, how many wooden objects would be there in the image?\\ & taxonomic & If one person leaves the cleaning group, how many mammals would remain?\\ & functional & If the two bedside lamps were removed, how many objects are present that need electricity?\\ & relational & If the signages were removed, how many objects would be present that hang from the ceiling?\\
\bottomrule
\end{tabular}
% } %
\end{table*}

\begin{figure*}[!t]
    \centering
    
    % --- STEP 1: Measure the pie chart ---
    % We save the image into a hidden box (\sbox0) at the exact width it will be displayed (48% of the page).
    \sbox0{\includegraphics[width=0.38\linewidth]{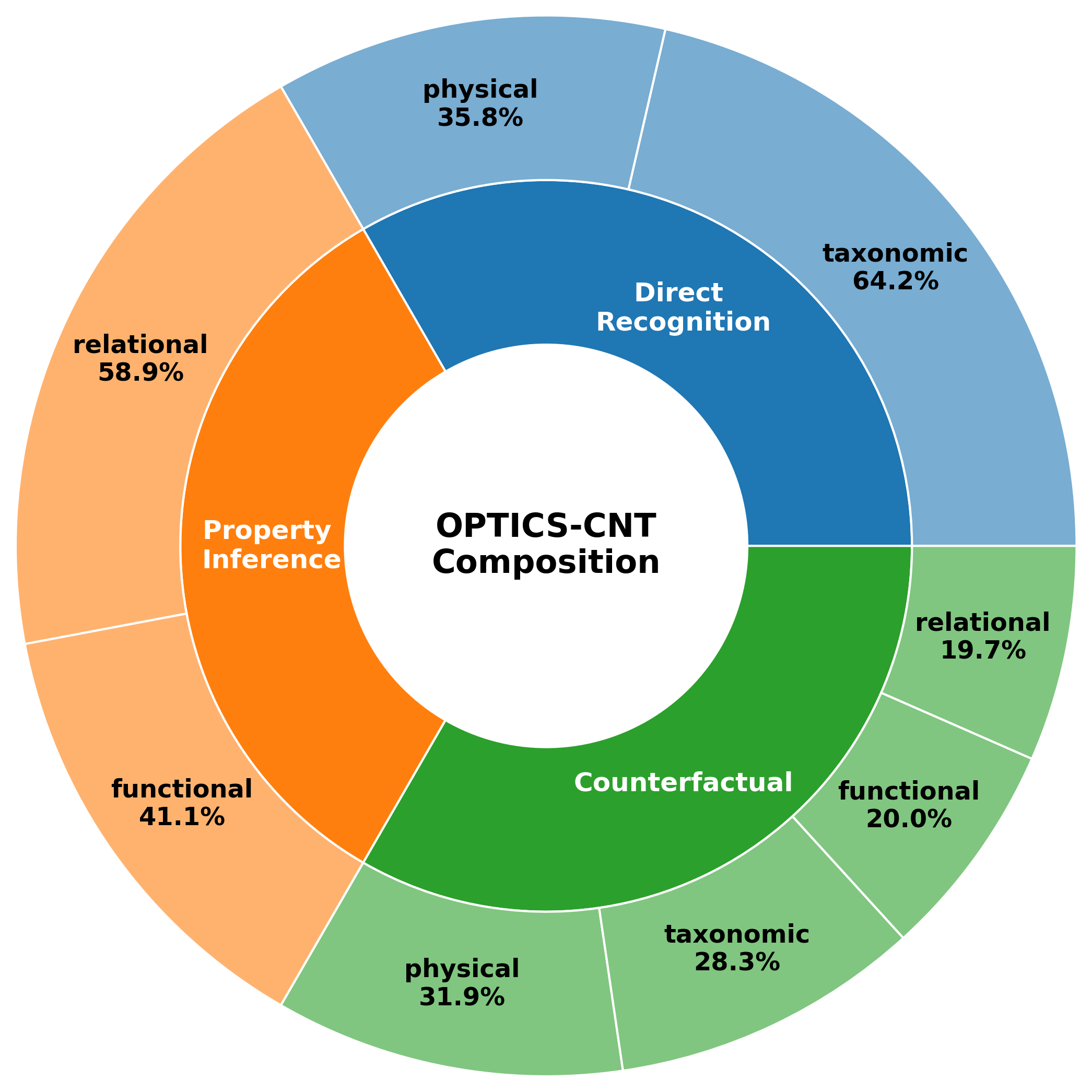}}
    
    % --- First Subfigure: tcolorbox ---
    \begin{subfigure}[b]{0.40\linewidth} % Keep [b] so captions align at the bottom!
        \centering
        % --- STEP 2: The Magic Wrapper ---
        % We create a minipage that is EXACTLY as tall as the pie chart (\ht0).
        % The [c] tells LaTeX to center the tcolorbox vertically inside this empty space.
        \begin{minipage}[b][\ht0][c]{\linewidth}
            \begin{tcolorbox}[
                colback=gray!5,
                colframe=gray!40,
                boxrule=0.4pt,
                arc=2pt,
                left=5pt,right=5pt,top=4pt,bottom=4pt,
            ]
            {\small
            Let $A$ be the answer to \lefttag{\textit{left question}} for the \lefttag{\textit{left image}}. 
            Let $B$ be the answer to \righttag{\textit{right question}} for the \righttag{\textit{right image}}. 
            Which is greater, $A$ or $B$?
            }
            \end{tcolorbox}
        \end{minipage}
        \caption{\second\ question template.}
        \label{fig:comparison_template}
    \end{subfigure}
    \hfill
    % --- Second Subfigure: Pie Chart ---
    \begin{subfigure}[b]{0.48\linewidth} % Keep [b] so captions align at the bottom!
        \centering
        % --- STEP 3: Print the measured image ---
        % \usebox0 simply prints the image we measured in Step 1.
        \usebox0
        \caption{\first\ question composition.}
        \label{fig:benchmark_compostion} 
    \end{subfigure}
    
    % --- Main Figure Caption and Label ---
    \caption{Overview of \benchmark\ question template (a) and composition split (b).}
    \label{fig:template_and_composition}
\end{figure*}

\noindent \textbf{Quality Assurance and Consolidation of \first.} Human annotators may introduce controversial assumptions and biases. To ensure broader consensus and address ambiguous questions, we conduct two rounds of quality assurance following the ``two annotators + third-annotator adjudication'' practice widely adopted in computer vision~\cite{biswas2025lecture} and NLP~\cite{li2024pedants}. First, each of the 360 images is randomly assigned to an annotator (from the same pool) who did not annotate the QA pair in the previous phase. This annotator is asked to answer the question based on the image. Comparing the two annotators' answers for each question, we observed that they agreed on 756 answers (70\%). We identified two key disagreement sources: 1) Semantic ambiguity: uncertainty in the question's criteria, where it is unclear whether an object that only partially satisfies a condition (e.g., a knife with a metal blade but a wooden handle) or an object with a broad definition (e.g., a container) should be counted.
2) Visual ambiguity: uncertainty in the visual information, such as deciding whether to count a partially visible chair.
% Moreover, we noticed that the difference in the numeric answers followed a Zipfian distribution for each of the three image types, with most disagreements differing by one (see Appendix). 
We addressed these disagreements with another round of quality assurance, where a third, different annotator resolved the disagreements based on the question, answers, and comments, by reformulating the questions to make them more precise, or if that was challenging, to replace them with a new QA pair from the same reasoning complexity level. % as the original question.
\textit{This results in \first.}

\begin{figure}[ht!] % [h!] is a placement specifier: here, top, bottom, page of its own
    \centering 
    \includegraphics[width=\linewidth]{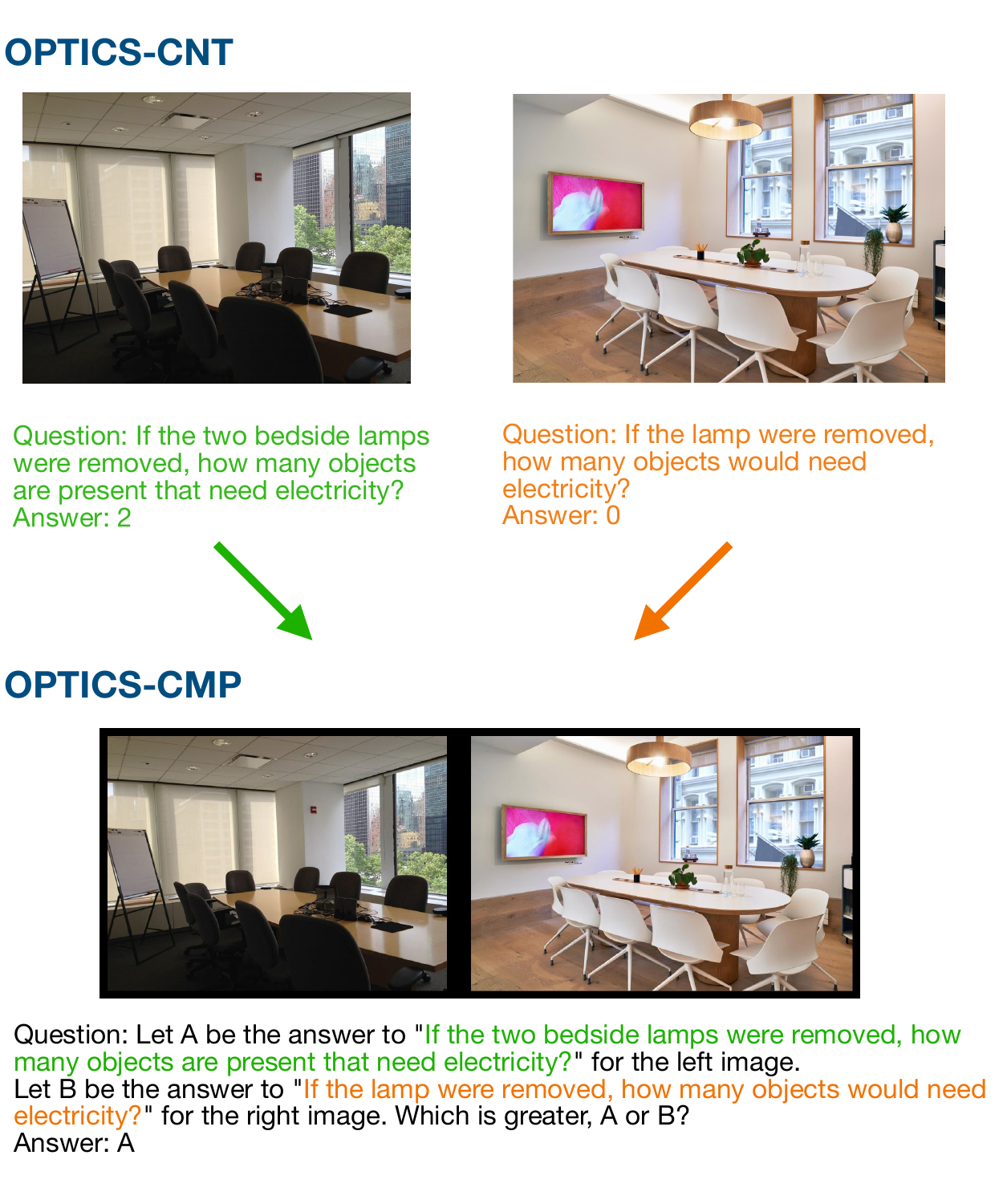} 
    \caption{Example from \second\ with a count difference of 2.}  % Caption for the image
    \label{fig:orbit_compare_example} 
\end{figure}

\noindent \textbf{Construction of \second.}  
% To introduce an evaluation axis beyond exact counting and diversify question types, we devise \second, which assesses the ability to make \textit{relative} judgments between images. 
To move beyond exact counts and diversify question types, we developed \second~to evaluate the ability to make relative judgments between images.
\second\ is derived from annotated VQA samples in \first. We first group questions by object properties and reasoning complexity, then pair them based on semantic similarity~\cite{reimers-2019-sentence-bert}. For each pair, the corresponding images are resized and concatenated. Then, a comparison question is generated using the template in \cref{fig:comparison_template}, and an answer is automatically derived from the \first\ labels (see example in \cref{fig:orbit_compare_example}).
We design \second\ questions to be fully compositional, inspired by metamorphic
testing~\cite{yuan2021perception}. This structure allows us to test whether VLMs
perform comparisons internally, verify consistency between \first~and \second~answers, and analyze uncertainty quantification and model attention (cf. §\ref{sec:results}).

\noindent \textbf{Benchmark Composition and Analysis.} \first\ integrates the images, questions, and final ground truth answers in a dataset containing 1,080 questions about 360 high-quality images of photographic, animated, and AI-generated types in equal proportions. 
% The distribution of questions across property dimensions and reasoning levels is depicted in Figure \cref{fig:benchmark_compostion}). 
 % visualizes the distribution of questions in \first\, indicating that %on the property of the object and the dimensions of the complexity of reasoning.
The overall distribution across object properties is %also 
relatively balanced (\cref{fig:benchmark_compostion}).
We observe a larger share of taxonomic than physical questions in the direct recognition category, and, similarly, slightly more relational than functional questions in the inference category. Curiously, when creating counterfactual questions,  annotators prioritized physical and taxonomic over relational and functional knowledge. \Cref{tab:reasoningcompelxity_questions} shows examples of \first\  questions at different reasoning levels.
Meanwhile, \second\ comprises 2,122 comparison questions generated from merged image pairs, with the count difference ranging from 1 to 9. Notably, 80\% of the \second~comparisons fall within a difference of 1–4, which is mathematically expected.\footnote{Small differences are combinatorially more numerous in the bounded range [0, 10] (diff=1 yields 10 pairs vs. only 2 for diff=9).} Regardless, our analysis is unaffected, as we report per-difference results across the full range.
The most common question pairs in \second\ involve direct recognition of an object's taxonomic~(56.7\%) or physical~(22.1\%) properties~(see \cref{fig:orbit_compare_statistic} for details).

% We observe that \textbf{TODO after we change the image}.
We finally estimated the \first\ difficulty for humans by asking two new participants to answer 90 questions about 30 random images from \first, split across reasoning levels and object properties. Their overall accuracy of 74\% %and an annotator agreement of 68\% 
indicates that while the task is challenging for humans, they answer most questions correctly.

%\textbf{R1} consists of 58% \textbf{D1} and 42% \textbf{D2} questions; \textbf{R2} comprises 92% \textbf{D3} and 8% \textbf{D4} questions; and \textbf{R3} features a balanced distribution of 24% \textbf{D1}, 22% \textbf{D2}, 30% \textbf{D3}, and 24% \textbf{D4} questions. Ground truth count distributions (0-10) are provided in Appendix \cref{sec: GT_distribution}. Table \cref{tab:benchmark_comparison} offers a comprehensive comparison of \benchmark with existing VQA benchmarks, highlighting its unique scope.

\section{Experimental Setup}% (0.5)}
% We here detail our experimental evaluation of several state-of-the-art Vision Language Models (VLMs) on the \benchmark\ benchmark. 

\noindent \textbf{Model Selection.}
% \label{ssec: model_selection}
% We include one closed-source and several open-source state-of-the-art Vision Language Models (VLMs), encompassing both instruction-tuned and non-instruction-tuned variants.
To ensure representative evaluation, we include a variety of non-instruction-tuned and instruction-tuned open-source VLMs, as well as a fine-tuned VLM and a closed-source model.  
As non-instruction-tuned open-source VLMs, we include BLIP-2 \cite{li2023blip} combined with OPT-2.7b and OPT-6.7b \cite{zhang2022opt}, and Fuyu-8b \cite{bavishi2023fuyu}. We include the instruction-tuned open-source models BLIP-2 paired with Flan T5-xxl \cite{chung2024scaling}, Qwen2.5-VL (3B, 7B, and 32B Instruct models) \cite{qwen2.5-VL}, and InternVL3 family (8B and 14B variants) \cite{zhu2025internvl3exploringadvancedtraining} prominent on the Hugging Face OpenVLM Leaderboard \cite{duan2024vlmevalkit}, and Gemma 3 \cite{gemmateam2025gemma3technicalreport} with 27B parameters. 
% We evaluate a specialized reasoning model that enhances specific reasoning capabilities. Namely, 
We incorporate the SpatialVLM fine-tuned version of Qwen2.5-VL for spatial relations and depth estimation in VQA \cite{Chen_2024_CVPR}, SpaceThinker-Qwen2.5-VL,
% , a model implemented by an open-source tool VQASynth \cite{VQASynth}, and 
claimed by the authors to be their most accurate specialized reasoning model. 
As a closed-source VLM, we include only GPT-4o-mini due to budget considerations \cite{openai2024gpt4ocard}.
% Finally, we include a baseline that samples a valid answer (0-10) uniform randomly, i.e., at 9.09\% accuracy. 
All models are evaluated in a zero-shot setting using standard hyperparameters.
%., a model trained specifically . 
% We also inlclude the open-source VQASynth \cite{VQASynth} combined with SpaceThinker-Qwen2.5-VL (3B) as its authors identified it as their most accurate fine-tuned model. 

% This selection enables direct comparison with the untuned Qwen2.5-VL-Instruct models, especially given their notably poor performance on relational properties. 
% For a fair comparison, we also evaluate the 3B parameter version of Qwen2.5-VL-Instruct \cite{qwen2.5-VL}.
 % Finally, we include a baseline that samples a valid answers (0-10) uniform randomly, i.e., at 9.09\% accuracy.
%Models are downloaded in their default float32 precision but loaded in float16 for experimental efficiency. We optimize GPU memory usage by directly loading model weights onto the device (low_cpu_mem_usage) and allow custom code execution from Hugging Face repositories (trust_remote_source_code). Output token lengths are adjusted per model, ranging from 30 for Fuyu 8b to 200 for Gemma 3. Our evaluations are conducted on the DAS-6 cluster \cite{bal2016medium}, utilizing SLURM job scripts across NVIDIA RTX A5000, NVIDIA A100, and NVIDIA RTX A6000 GPU nodes.
% All models are evaluated in a zero-shot setting using standard hyperparameters.

\noindent \textbf{Answer Extraction.} 
% To extract numeric answers from the model outputs for \first, we employ regular expressions for detection and perform exact string matching against ground-truth labels. Prompts were designed to facilitate this extraction by requesting only the total count (numeric) or a count followed by object lists (see Table \cref{tab:output_prompts}). The regular expression extraction is kept flexible to obtain counts provided in textual or Roman numeral format (observed with some non-instruction-tuned models). Counts greater than ten are uninformative in the analysis of deviation from the ground-truth and therefore are considered invalid. We interpret missing outputs as a prediction of zero. To extract the answers for \second, we designed the prompt so that A refers to the counting question about the left image and B refers to the counting question about the right image. We instructed the model to output only “A” or “B”, indicating which value is greater, without giving any explanation. We also used regular expressions to detect the correct answer in cases where the model still produced extra text instead of giving the direct letter.
For both benchmarks, our VLM prompts facilitate answer extraction by requesting either the total count or a count followed by objects for \first\ (see \cref{tab:output_prompts}), and only ``A'' or ``B'' to indicate which value is greater for \second.
To extract answers from \first\ (numeric label) and \second\ (A or B label), we employ flexible regular expressions for detection and perform exact string matching against ground-truth labels.
% The regular expression extraction is kept flexible to obtain counts provided in textual or Roman numeral format (observed with some non-instruction-tuned models) and the A or B label from outputs containing extra text. 
For \first, we interpret missing outputs as a prediction of zero, and though during data curation we restrict object counts to 10 per image, we allow model outputs with counts greater than 10 to analyze deviation from ground truth. 
% \noindent \textbf{Open-source VLMs. } For the \textbf{non-instruction-tuned} models, we test 1) BLIP-2 (\cite{li2023blip}), paired with the OPT-2.7b and the OPT-6.7b language models (\cite{zhang2022opt}), where BLIP2 handles the visual part and the OPT variants the text generation, forming a vision-language pipeline; 2) Fuyu-8b (\cite{bavishi2023fuyu}), a multimodal text and image transformer that achieves competitive performance on VQA tasks.

\noindent \textbf{Evaluation Metrics.}
Following established count-based VQA benchmarks~\cite{trott2018interpretablecountingvisualquestion,acharya2018tallyqaansweringcomplexcounting},
% like HowManyQA \cite{trott2018interpretablecountingvisualquestion} and TallyQA \cite{acharya2018tallyqaansweringcomplexcounting}, 
we report multiple evaluation metrics for \first. Our primary measure is micro \textit{accuracy (Acc)}. We report \textit{macro accuracy}, which averages the accuracy scores across different counts, ranging from 0 to 10.
We also report \textit{Root-Mean-Squared Error (RMSE)} to quantify the typical deviation between predicted and ground-truth counts. To diagnose model bias, we include \textit{mean error}, which reveals bias towards over- or under-counting. Furthermore, \textit{off-by-N} accuracy (cumulative accuracy by tolerance) indicates the proximity of the predictions to the correct count.
We also report accuracy per count difference for \second, 
% the overall micro and macro accuracy and 
% % macro-accuracy 
% across count differences, 
expecting an increase in accuracy for larger differences.
% increase proportionally with the count difference increases.

\noindent \textbf{Uncertainty Analysis.}
We conduct an uncertainty analysis to investigate the models' confidence in their predictions, and to understand the degree to which such scores distinguish correct answers from incorrect ones on \second. Following recent conventions~\cite{wang2026vlm, huang2025visual}, we first assign labels to all predictions: \texttt{0} if correct, and \texttt{1} otherwise. Then, we compute an uncertainty score from the model’s predicted probability scores using several uncertainty estimators, which are detailed below. We then use these estimators to rank predictions in ascending order, i.e., with predictions with high model confidence first. To assess uncertainty calibration, we finally evaluate this ranking using AUROC, treating incorrect predictions as the positive class with label \texttt{1}, to indicate that higher uncertainty should correspond to a higher probability of error.

The included uncertainty estimators jointly provide a computationally efficient uncertainty quantification: (i) maximum sequence probability (MSP) extends softmax probabilities to the autoregressive setting, (ii) perplexity (PPL) is defined as the exponentiated average negative log-likelihood of the sequence, and (iii) mean token entropy (MTE) captures the dispersion across the entire token distribution rather than only the generated tokens. Estimates of metrics were used where necessary, e.g., for black-box GPT models~\cite{shelmanovvashurin2025}.

\begin{table*}[!t]
\caption{Zero-shot results on \first. % The zero-shot (micro) accuracy, macro accuracy, RMSE, mean error, and off-by-1 accuracy are reported.
\textbf{VLMs exceed the random baseline but fall short of human accuracy and exhibit undercounting bias, with performance generally scaling with model size and instruction tuning.}
The highest and second-highest model results are highlighted in \textbf{bold} and \underline{underlined}, respectively. ($\uparrow$)/($\downarrow$) indicates that higher/lower values are better. ($\rightarrow0$) signifies that values closer to zero are better. $^{\dagger}$ means the result is attributed to inductive bias.}
% $\dagger$ notes that the result is attributed to inductive bias.}
\label{tab:model_results}
\centering
\small
% Adjust column separation. Default is usually 1pt. Try a larger value.
% \setlength{\tabcolsep}{2pt}
% \resizebox{0.75\textwidth}{!}{%
% \makebox[\linewidth][c]{%
\begin{tabular}{l|r|r|r|r|r}
\toprule
% \textbf{Category} & \textbf{Model} & \textbf{AVR Question} & \textbf{Fine-grained Perception} & \textbf{Perc C} & \textbf{Perc C\&F} & \textbf{Perc C\&F \& AVR} \\
% \multirow{1}{*}{
% \textbf{Category}} & 
\textbf{Model} & 
\parbox{2cm}{\hfill \textbf{Micro Acc ($\uparrow$)}}
& \parbox{2.1cm}{\hfill \textbf{Macro Acc ($\uparrow$)}} &
% \textbf{Acc($\uparrow$)} & 
\parbox{1.4cm}{\hfill \textbf{RMSE ($\downarrow$)}} & \parbox{2.5cm}{\hfill \textbf{Mean Error ($\rightarrow0$)}} & \parbox{1.8cm}{\hfill \textbf{Off-By-1 ($\uparrow$)}} \\
\midrule
% \multirow{1}{*}{\textbf{-}} &
\textit{Random} & 9.09 & 9.09 & 4.47 & 0.00 & 25.62 \\
\midrule
% \multirow{9}{*}{\parbox{1.5cm}{\centering \textbf{Open-source}}}
% & 
BLIP-2 OPT (2.7B) & 12.87$^{\dagger}$ & 10.92 & 3.66 & -2.68 & 35.28 \\
% & 
BLIP-2 OPT (6.7B) & 16.67$^{\dagger}$ & 9.68 & 3.06 & -1.56 & 46.11\\
% & 
Fuyu (8B) & 19.91 & 11.88 & 2.68 & -0.89 & 49.72 \\
% \cmidrule{2-5}
\midrule
% & 
BLIP-2 Flan (11B) & 15.93$^{\dagger}$ & 12.95 & 3.53 & -2.05 & 42.59 \\
% & 
Qwen2.5 (3B) & 24.91 & 23.55 & 15.30 & -1.16 & 51.85 \\
% & 
Qwen2.5 (7B) & 38.70 & \underline{31.57} & \underline{1.88} & -0.62 & \underline{70.28} \\
% & 
Qwen2.5 (32B) & \textbf{39.91} & \textbf{34.46} & \textbf{1.80} & -0.35 & \textbf{72.78} \\
% & 
SpaceThinker (3B) & 29.63 & 25.32 & 2.19 & -0.89 & 62.50\\
% & 
InternVL3 (8B) & 37.59 & 31.30 & 1.93 & -0.70 & \underline{70.28}\\
% & 
InternVL3 (14B) & \underline{39.72} & 29.90 & 2.08 & -0.91 & 68.61\\
% & 
Gemma3 (27B) & 31.76 & 24.43 & 2.20 & -0.14 & 65.28\\
\midrule
% \textbf{Closed-source}
% & 
GPT-4o mini & 30.37 & 22.74 & 15.19 & -0.36 & 64.17\\
\midrule
% \multirow{1}{*}
\textit{Model average} & 28.16 & 22.39 & 4.62 & -1.02 & 58.28 \\ \midrule
\textit{Human 1} & 75.56 & 75.15 & 4.22 & 0.64 & 82.22\\
\textit{Human 2} & 72.22 & 67.27 & 1.63 & -0.24 & 85.56\\
\textit{Human average} & 73.89 & 71.21 & 2.92 & 0.2 & 83.89 \\
% & Gemini-pro-vision* & 25.00$^{\dagger}$ & 44.42 & 15.19 & 6.75 \\
% & Claude3 (Sonnet) & 26.49$^{\dagger}$ & 50.91 & 38.70 & 19.87 \\
% & Claude3 (Opus) & \textbf{28.83} & 47.27 & \textbf{44.94} & \textbf{20.13} \\
% \midrule
% & Human & $68.86 \pm 9.74$ & - & - & - & - \\
\bottomrule
\end{tabular}
% }
% } %
\end{table*}

\section{Results}% (1.5)}
\label{sec:results}

% We investigate three research questions: 1) How well can VLMs accurately count and reason about object properties? 2) How does the performance of VLMs vary across the different property dimensions, reasoning levels, and image types? and 3) Do VLMs benefit from property-specific fine-tuning?

\textbf{How well can VLMs accurately count and reason about object properties?}
% \label{ssec: overall_performance}
The results for \first\ in \cref{tab:model_results} show that 
% for both the open- and closed-source categories, 
\textit{most models perform above random, but well below human accuracy (28\% versus 74\% on average).} The high RMSE of several models indicates 
% substantial variance in predictions, suggesting 
that precise numerical grounding is challenging, even for strong VLMs. The negative mean error values suggest consistent underestimation and a general bias towards undercounting.
% the more negative the mean error values, the lower the model calibration and prediction stability. 
We note a severely imbalanced prediction distribution in the outputs of some VLMs (detailed in \cref{fig:inductive_bias}). When allowing for one counting error, the accuracy increases to 58\% on average (to 73\% for the best model), which indicates that many model errors are near-misses (see \cref{fig:cumulative_accuracies} for an in-depth analysis of the top-3 best performing VLMs). For all VLMs, the micro accuracy is higher than the macro accuracy because all models exhibit a bias towards lower, more frequent counts, as shown in \cref{fig:accuracy_per_count}. In contrast, human accuracy is uniform across answer counts, with no apparent frequency bias.
We further analyze the sources of errors for both humans and VLMs at the end of this section.

\noindent \textbf{How does counting performance vary across VLMs?}
Accuracies vary widely across models, ranging from 13\% to 40\%. %, between % is , ranging from % exhibits a broad spread from 
% 12.87\% and 39.91\%. % across the 12 models. 
% Some models show limited improvements, while others show moderate gains. 
% Among the VLMs, 
The open-source Qwen2.5-VL-Instruct (32B) is the best-performing model, with the highest exact and off-by-1 accuracies, the lowest RMSE, and the lowest mean error. A small gap separates it from InternVL3 (14B), with the second-highest overall accuracy, and Gemma 3 (27B), which achieves the lowest mean error. 
% The second-best off-by-1 accuracy is tied between Qwen2.5-VL-Instruct (7B) and InternVL3 (8B). 
The closed-source GPT-4o-mini exhibits moderate performance, with a high RMSE and low mean error. 
% Additionally, the analysis indicates that the predictions are generally stable (Table \cref{tab:model_results}). 
We also note a typical increase in performance with increasing model size within model families, though this may not always hold across architectures. \Cref{tab:dimension_analysis} shows that Qwen2.5-VL-Instruct (7B) and (32B) perform the best across nearly all reasoning levels, property dimensions, and image types, while GPT-4o-mini performs consistently worse than most instruction-tuned open-source models.

\begin{figure}[ht]
    \centering
  \resizebox{\linewidth}{!}{\begin{tikzpicture}[scale=0.8, baseline=(current axis.north)]
\begin{axis}[
    width=5in,
    height=3in,
    xlabel={Ground Truth Count},
    ylabel={Accuracy (\%)},
    title={},
    xmin=-0.2, xmax=10.2,
    ymin=0, ymax=100,
    xtick={0,1,2,3,4,5,6,7,8,9,10},
    ytick={0,20,40,60,80,100},
    grid=major,
    grid style={gray!20, line width=0.5pt},
    % legend pos=north east,
    legend style={font=\scriptsize, legend cell align=left, fill opacity=.7, at={(0.5, 1.3)}, anchor=north},
    legend columns=4,
    every axis plot/.append style={line width=1pt, mark size=2.5pt},
    tick label style={font=\footnotesize},
    label style={font=\small},
    tick style={line width=0.6pt},
    x tick label style={/pgf/number format/fixed},
    y tick label style={/pgf/number format/fixed},
]

\addplot[color=purple!80!black, mark=diamond, mark options={fill=purple!80!black}] coordinates {
    (1, 94.4)
    (2, 79.4)
    (3, 82.1)
    (4, 44.4)
    (5, 75.0)
    (6, 72.7)
    (7, 37.5)
    (8, 71.4)
    (9, 75.0)
    (10, 80.0)
};
\addlegendentry{Human};

\addplot[color=red!80!black, mark=square, mark options={fill=red!80!black}] coordinates {
    (0, 56.5)
    (1, 68.5)
    (2, 69.7)
    (3, 49.0)
    (4, 33.7)
    (5, 13.8)
    (6, 18.9)
    (7, 4.4)
    (8, 2.0)
    (9, 8.8)
    (10, 3.4)
};
\addlegendentry{Intern14B};

\addplot[color=green!70!black, mark=triangle, mark options={fill=green!70!black}] coordinates {
    (0, 65.2)
    (1, 48.4)
    (2, 61.6)
    (3, 22.4)
    (4, 65.7)
    (5, 31.7)
    (6, 12.2)
    (7, 8.9)
    (8, 8.0)
    (9, 5.9)
    (10, 17.2)
};
\addlegendentry{Qwen7B};
\addplot[color=blue!80!black, mark=o, mark options={fill=blue!80!black}] coordinates {
    (0, 87.0)
    (1, 51.6)
    (2, 61.1)
    (3, 43.8)
    (4, 37.8)
    (5, 27.6)
    (6, 28.9)
    (7, 6.7)
    (8, 16.0)
    (9, 11.8)
    (10, 6.9)
};
\addlegendentry{Qwen32B};

\end{axis}
\end{tikzpicture}}
    \caption{Accuracy per count (0-10) on \first\ for human with an average (micro) accuracy of 73.89\% and the top-3 best performing models: Qwen2.5-VL-Instruct (32B), InternVL3 (14B), and Qwen2.5-VL-Instruct (7B).}
    \label{fig:accuracy_per_count}
\end{figure}
  % \hfill
  
\begin{table*}[!t]
\caption{Accuracy results on \first\ across reasoning levels (\underline{D}irect \underline{R}ecognition, \underline{P}roperty \underline{I}nference, \underline{C}ounterfactual \underline{R}easoning) 
object property dimensions (\underline{P}hysical, \underline{T}axonomic, \underline{F}unctional, \underline{R}elational), 
% [D1: Physical, D2: Taxonomic, D3: Functional, D4: Relational]; 
and image types.
\textbf{Qwen2.5 and InternVL3 dominate across reasoning and property levels, though all models struggle with counterfactual logic and the complexity of photographic images compared to animated or AI-generated scenes.}
% [I1: Real, I2: Animated, I3: AI-Generated]. 
The highest and second-highest model results are highlighted in \textbf{bold} and \underline{underlined}, respectively.}
\label{tab:dimension_analysis}
\centering
\small
% Adjust column separation. Default is usually 1pt. Try a larger value.
% \setlength{\tabcolsep}{2pt}
% \resizebox{\textwidth}{!}{%
% \makebox[\linewidth][c]{%
\begin{tabular}{l|rrr|rrrr|rrr}
\toprule
\textbf{Model} 
& \multicolumn{3}{c|}{\textbf{Reasoning Complexity}} 
& \multicolumn{4}{c|}{\textbf{Object Properties}} 
& \multicolumn{3}{c}{\textbf{Image Types}} \\
% \cmidrule(lr){2-4} \cmidrule(lr){5-8} \cmidrule(lr){9-11}
\cmidrule{2-4} \cmidrule{5-8} \cmidrule{9-11}
& DR & PI & CR & P & T & F & R 
& Photo & Animated & AI Gen. \\
\midrule
BLIP-2 OPT (2.7B) & 10.83 & 13.06 & 14.72 & 11.48 & 14.41 & 10.91 & 13.78 & 9.17 & 19.44 & 10.00 \\
BLIP-2 OPT (6.7B) & 16.67 & 18.06 & 15.28 & 16.39 & 16.52 & 14.09 & 19.08 & 14.17 & 21.11 & 14.72 \\
Fuyu (8B) & 18.33 & 22.78 & 18.61 & 17.21 & 17.42 & 19.55 & 25.44 & 16.94 & 23.89 & 18.89 \\
\midrule
BLIP-2 Flan (11B) & 18.33 & 18.61 & 10.83 & 14.34 & 16.82 & 13.64 & 18.02 & 11.67 & 20.00 & 16.11 \\
Qwen2.5 (3B) & 26.11 & 29.17 & 19.44 & 21.31 & 29.43 & 17.73 & 28.27 & 20.28 & 27.78 & 26.67\\
Qwen2.5 (7B) & \textbf{42.78} & \textbf{44.17} & 29.17 & \textbf{35.66} & 42.34 & 32.73 & \underline{41.70} & \textbf{32.22} & 40.83 & \textbf{43.06} \\
Qwen2.5 (32B) & \underline{39.17} & \underline{43.06} & \textbf{37.50} & \underline{33.61} & \underline{44.44} & \textbf{38.18} & 41.34 & \underline{31.94} & \underline{45.00} & \underline{42.78} \\
SpaceThinker (3B) & 34.44 & 30.28 & 24.17 & 25.00 & 37.24 & 19.55 & 32.51 & 25.56 & 31.11 & 32.22 \\
InternVL3 (8B) & 38.89 & 41.94 & 31.94 & 33.20 & 42.34 & 26.82 & \textbf{44.17} & 28.89 & 43.61 & 40.28 \\
InternVL3 (14B) & \textbf{42.78} & 42.22 & \underline{34.17} & 30.33 & \textbf{48.05} & \underline{36.82} & 40.28 & \textbf{32.22} & \textbf{48.61} & 38.33 \\
Gemma3 (27B) & 33.89 & 35.56 & 25.83 & 26.23 & 34.53 & 28.64 & 35.69 & 27.78 & 36.39 & 31.11 \\
\midrule
GPT-4o mini & 28.89 & 33.61 & 28.61 & 21.31 & 34.23 & 25.00 & 37.81 & 24.44 & 38.89 & 27.78 \\ 
\midrule
\textit{Model average} & 29.34 & 31.04 & 24.18 & 23.83 & 31.48 & 23.63 & 31.50 & 22.94 & 33.05 & 28.49\\
\midrule
\textit{Human 1} & 73.33 & 80.00 & 73.33 & 58.33 & 92.00 & 71.43 & 80.00 & 63.33 & 83.33 & 80.00 \\
\textit{Human 2} & 66.67 & 76.67 & 73.33 & 58.33 & 80.00 & 76.19 & 75.00 & 60.00 & 83.33 & 73.33 \\
\textit{Human average} & 70.00 & 78.33 & 73.33 & 58.33 & 86.00 & 73.81 & 77.50 & 61.66 & 83.33 & 76.66\\
\bottomrule
\end{tabular}
% } %
\end{table*}

%   % --- Main Figure Caption and Label ---
%   \caption{Line plots showcasing performance on per count and count difference for \first\ and \second\, respectively. \textbf{(a)} Accuracy per count (0-10) on \first\ for human with an average (micro) accuracy of 73.89\% and the top-3 best performing models: Qwen2.5-VL-Instruct (32B), InternVL3 (14B), and Qwen2.5-VL-Instruct (7B). \textbf{(b)} Accuracy per count difference (1-9) for instruction-tuned VLMs on \second. Counts (6-9) are grouped as one.}
%   \label{fig:line_plots}
% \end{figure*}

\noindent \textbf{How does counting performance vary across reasoning levels?}
% \label{ssec: dimension_based_analysis}
% Based on the presented framework, 
% Regarding the reasoning levels, as discussed in Section \cref{ssec: reasoning_levels}, 
We expect the progression of reasoning complexity \textit{recognition $<$ property inference $<$ counterfactual} (\cref{sec:framework}) to be reflected in corresponding trends in model performance. However, the results among the top-tier VLMs (Qwen2.5-VL-Instruct, InternVL3, Gemma 3, and GPT-4o-mini) reveal a progression \textit{property inference $<$  recognition $<$ counterfactual}. This suggests that modern VLMs still exhibit poor low-level perceptual capabilities, which aligns 
% likely because they are trained on descriptive and semantically rich corpora that emphasize abstract associations over precise recognition. These perceptual limitations align 
with several recent findings: 1) 
% First, 
Jiang \etal \cite{jiang2024marvel} showed poor perceptual performance of MLLMs for fine-grained visual analysis tasks, including counting;
%and .
% Second, 
2) Zhang \etal \cite{zhang2024exploring} identified blind spots in MLLMs, including sensitivity to visual quality and image distractors; %And, finally, 
% 3) Yuan \etal \cite{yuan2021perception} showed perception failures in the MetaVQA framework and highlighted the need for detailed perceptual grounding of VLMs. 
3) Zhang \etal \cite{zhang2025sphere} revealed spatial blind spots in VLMs through a hierarchical evaluation framework.
The lower accuracy on counterfactual questions is indicative of difficulty with changes and hypotheticals, reliance on learned associations, and lack of reliable grounding~\cite{li2024eyes}.
% Similarly to VLMs, 
Humans also perform best on inference questions but struggle less than VLMs with counterfactual reasoning.

\noindent \textbf{How does counting accuracy vary across object properties?}
Among the object property dimensions, most models (as well as humans) have a stronger grasp of taxonomic and relational questions than of physical and functional ones. 
This finding is in line with recent work that points out the struggle of LLMs
% struggle of models 
with physical object attributes~\cite{wang2023newton}, physical interactions~\cite{aroca2021prost}, functional requirements~\cite{qasemi-etal-2022-paco}, and atypical affordances~\cite{tian2025macgyverlargelanguagemodels}. 
% This could be because 
% physical and functional dimensions require more hops of reasoning as they are usually more indirect and difficult to discern (similar trend with humans).
% and the possibility of more number of lower count questions in the taxonomic and relational dimensions, if answered correctly boosting their performance.
% Two reasons for this improved performance could be more number of questions for the taxonomic and relational dimensions and the possible data leakage in questions where objects are addressed directly by their nouns instead of a descriptive phrase, for example, (\textit{an object used for transport}, \textit{a red item}) instead of (\textit{a car}, \textit{an apple}), potentially preventing the previous reason. 
We note an improvement for the fine-tuned SpaceThinker-Qwen2.5-VL (3B) model compared to its non-fine-tuned counterpart, Qwen2.5-VL (3B), across the four object property dimensions. 
This shows that, in addition to relational questions, other object properties also benefit from spatial tuning of Qwen2.5 (3B). %to perform better not only on relational questions (4.24\%), but also on other dimensions that may require similar object property knowledge. %partially hinge on similar reasoning. 
% possibly due to underlying spatial and depth aspects involved while locating objects in an image, allowing scope for improvements in the remaining dimensions through property specific fine-tuning.
We also observe that the InternVL3 models have a better grasp of taxonomic and relational questions compared to other models. The Qwen2.5-VL (7B) and (32B) variants demonstrate consistent performance across all properties, with particular strength in physical and functional questions. In contrast, the non-instruction-tuned models struggle across all properties, underscoring their weakness in visual object reasoning.

\begin{figure}[ht]
    \centering
    \resizebox{\linewidth}{!}{\begin{tikzpicture}[scale=0.75, baseline=(current axis.north)]
\begin{axis}[
    width= 5in,
    height=3in,
    xlabel={Ground Truth Count Difference},
    ylabel={Accuracy (\%)},
    xmin=-0.5, xmax=5.5,
    ymin=40, ymax=80,
    xtick={0,1,2,3,4,5},
    xticklabels={1,2,3,4,5,6-9},
    ytick={40,50,60,70,80},
    grid=major,
    grid style={gray!20, line width=0.5pt},
    % legend pos=north west,
    legend style={font=\scriptsize, legend cell align=left, fill opacity=.7, at={(0.5, 1.32)}, anchor=north},
    legend columns=3,
    every axis plot/.append style={line width=1pt, mark size=2.5pt},
    tick label style={font=\footnotesize},
    label style={font=\small},
    tick style={line width=0.6pt},
    x tick label style={/pgf/number format/fixed},
    y tick label style={/pgf/number format/fixed},
]
\addplot[color=red!80!black, mark=o, mark options={fill=red!80!black}] coordinates {
    (0, 53.2)
    (1, 59.9)
    (2, 60.2)
    (3, 58.6)
    (4, 54.9)
    (5, 70.0)
};
% \addlegendentry{gemma-3-27b-it};
\addlegendentry{Gemma27B};
\addplot[color=blue!80!black, mark=square, mark options={fill=blue!80!black}] coordinates {
    (0, 53.6)
    (1, 59.6)
    (2, 60.3)
    (3, 62.3)
    (4, 69.2)
    (5, 70.6)
};
% \addlegendentry{gpt-4o-mini};
\addlegendentry{Gpt4omini};
\addplot[color=green!70!black, mark=triangle, mark options={fill=green!70!black}] coordinates {
    (0, 58.6)
    (1, 63.1)
    (2, 60.0)
    (3, 67.0)
    (4, 53.0)
    (5, 67.9)
};
% \addlegendentry{internvl3-14b};
\addlegendentry{Intern14B};
\addplot[color=orange!80!black, mark=diamond, mark options={fill=orange!80!black}] coordinates {
    (0, 56.6)
    (1, 60.7)
    (2, 64.4)
    (3, 65.7)
    (4, 64.9)
    (5, 68.9)
};
% \addlegendentry{internvl3-8b};
\addlegendentry{Intern8B};
\addplot[color=purple!80!black, mark=asterisk, mark options={fill=purple!80!black}] coordinates {
    (0, 61.0)
    (1, 68.7)
    (2, 70.0)
    (3, 68.7)
    (4, 78.9)
    (5, 75.9)
};
% \addlegendentry{qwen2-5-vl-32b};
\addlegendentry{Qwen32B};
\addplot[color=teal!80!black, mark=pentagon, mark options={fill=teal!80!black}] coordinates {
    (0, 55.7)
    (1, 58.8)
    (2, 54.5)
    (3, 55.6)
    (4, 48.1)
    (5, 59.6)
};
% \addlegendentry{qwen2.5vl-3b};
\addlegendentry{Qwen3B};
\addplot[color=magenta!80!black, mark=x, mark options={fill=magenta!80!black}] coordinates {
    (0, 54.7)
    (1, 58.6)
    (2, 56.3)
    (3, 58.8)
    (4, 60.4)
    (5, 64.0)
};
% \addlegendentry{qwen2.5vl-7b};
\addlegendentry{Qwen7B};

\end{axis}
\end{tikzpicture}}
    \caption{Accuracy per count difference (1-9) for instruction-tuned VLMs on \second. Counts (6-9) are grouped as one.}
    \label{fig:accuracy_per_count_diff}
\end{figure}

% Regarding the image types, we see a general pattern of ease in answering AI-generated and animated images, but a challenge with real images. This is likely due to the visually cleaner and less cluttered AI-generated and animated image compared to the complex natural scenes from a real image. This performance trend aligns with previous claims made about synthetic data to real-world data by other benchmarks mentioned in section \cref{ssec: image_types}.

\noindent \textbf{How does counting performance vary across image types?}
We observe a consistently higher accuracy for AI-generated and animated images, compared to photographic images. We hypothesize that this is due to the visually cleaner, less cluttered nature of AI-generated and animated images compared to the complex, noisy real-world scenes in photographic images. This finding aligns with 
% those of 
previous studies reporting similar gaps between synthetic and real-world data \cite{johnson2017clevr, hudson2019gqa}. % that report similar gaps between synthetic and real-world data.
Moreover, we note that animated images are easier for most models than AI-generated images, possibly because the latter may have poor focus and include implausible objects, such as a half-cut glass floating in the air.
The trend between the image difficulty \textit{animated $<$ AI-generated $<$ photographic} is also clear in humans, which indicates that questions about photographic images may be the hardest and possibly contain remaining ambiguity. 
% To support the above analysis, a visual depiction
% Examples of model errors across image types are included in the Appendix. % \cref{s ec: appendix_errors}.

\begin{table}[h!]
\caption{Zero-shot results on \second. % The zero-shot (micro) accuracy, macro accuracy, RMSE, mean error, and off-by-1 accuracy are reported. 
The micro accuracy and macro accuracy (average across count differences (1-9)) are reported.
The highest and second-highest model results are highlighted in \textbf{bold} and \underline{underlined}, respectively. ($\uparrow$) indicates that higher values are better.}
% $\dagger$ notes that the result is attributed to inductive bias.}
\label{tab:model_results_compare}
\centering
\small
% Adjust column separation. Default is usually 1pt. Try a larger value.
\setlength{\tabcolsep}{1pt}
% \makebox[\linewidth][c]{%
\begin{tabular}{l|r|r}
\toprule
% \textbf{Category} & \textbf{Model} & \textbf{AVR Question} & \textbf{Fine-grained Perception} & \textbf{Perc C} & \textbf{Perc C\&F} & \textbf{Perc C\&F \& AVR} \\
% \multirow{1}{*}{
% \textbf{Category}} & 
\textbf{Model} & 
\parbox{2.1cm}{\raggedright \textbf{Micro Acc ($\uparrow$)}}
& \parbox{2.2cm}{\raggedright \textbf{Macro Acc ($\uparrow$)}} \\
% &
% \textbf{Acc($\uparrow$)} & 
% \parbox{0.9cm}{\centering \textbf{RMSE\\($\downarrow$)}} & \parbox{0.9cm}{\textbf{Mean\\Error\\($\downarrow$)}} & \parbox{0.7cm}{\textbf{Off-By-1\\($\uparrow$)}} \\
\midrule
Qwen2.5 (3B) & 55.14 & 56.30 \\
% & 
Qwen2.5 (7B) & 57.54 & 58.10  \\
% & 
Qwen2.5 (32B) & \textbf{68.33} & \textbf{72.39}  \\
% & 
% SpaceThinker (3B) & 29.63 & 25.32 & 2.19 & -0.89 & 62.50\\
% & 
InternVL3 (8B) & \underline{62.02} & 64.35 \\
% & 
InternVL3 (14B) & 46.84 & 64.19 \\
% & 
Gemma3 (27B) & 54.52 & \underline{64.48} \\
\midrule
% \textbf{Closed-source}
% & 
GPT-4o mini & 60.34 & 64.45 \\
\midrule
% \multirow{1}{*}
\textit{Model average} & 57.81 & 63.46 \\ 
\bottomrule
\end{tabular}
% }
\end{table}

% \Cref{tab:model_results_compare} shows the results for seven instruction-tuned models evaluated on \second. 
 
\textbf{How well do VLMs answer comparison questions, and is this sensitive to the count differences?} 
% \todo{add figure with accuracy per count diffs, plus add a paragraph with commentary in the paper}
Based on results from \first, we evaluate only the instruction-tuned models on \second. 
This experiment confirms that VLMs struggle with counting objects, even in this relative setting. Namely, VLMs' accuracy remains only slightly higher than random, at 58\% (\cref{tab:model_results_compare}). 
% show that models achieve modest improvements on relative comparison questions to absolute counting tasks (58\% vs 28\% micro accuracy on average). 
Qwen2.5-VL (32B) remains the overall best-performing model with micro accuracy of 68\%, followed by InternVL3 (8B), which outperforms its larger (14B) variant. This pattern does not hold within the Qwen2.5-VL family, where we instead observe a steady increase in performance as model size increases. 
% For all models, macro accuracy is higher than micro accuracy. 
Our hypothesis that VLMs would show higher gains on larger count differences is examined
% The relationship between performance and count difference is shown 
in \cref{fig:accuracy_per_count_diff}.
% shows models exhibit a bias towards higher, less frequent counts, which is opposite to the frequency bias observed in \first. 
% We hypothesized that ; 
While some models clearly follow this trend, we note that model performance across the count differences remains broadly stable.
% it is not consistent across all of them. For example, InternVL3 (14B) displays an erratic pattern. 
These findings confirm those from the absolute-counting task and suggest that the failure modes of humans and VLMs may differ.

\begin{table}[t!]
\caption{AUROC scores for three uncertainty estimates on \second. Higher values indicate that high uncertainty is more predictive of errors, MSP=maximum sequence probability, PPL=perplexity, MTE=mean token entropy.}
\label{tab:uncertainty}
\centering
% \small
% \resizebox{\linewidth}{!}{
\begin{tabular}{l|r|r|r}
\toprule
\textbf{} &
\textbf{MSP} &
\textbf{PPL} &
\textbf{MTE}  \\
\midrule
Qwen2.5 (7B)         & \textbf{0.64} & \textbf{0.66} & \textbf{0.64} \\
Qwen2.5 (32B)  &  \textbf{0.64} & 0.64 & 0.60 \\
GPT-4o mini  &  0.58 & 0.57 & - \\
\bottomrule
\end{tabular}
% }
\end{table}
\begin{figure}[t!] 
    \centering 
    \includegraphics[width=0.9\linewidth]{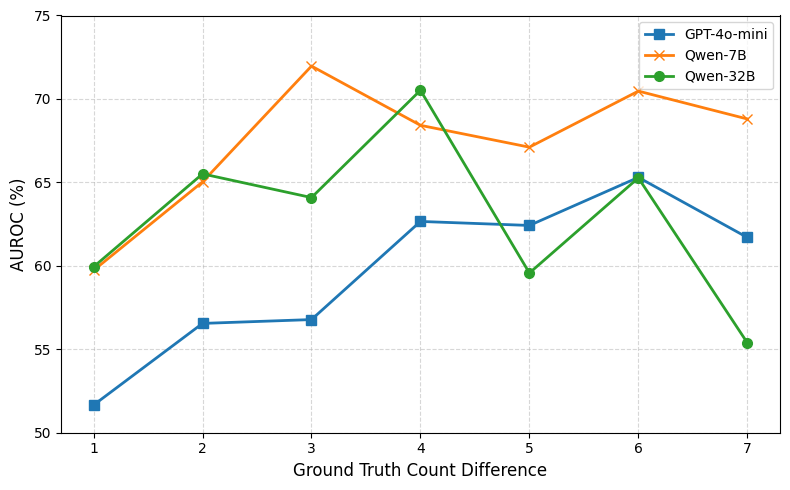} 
    \caption{AUROC for perplexity across ground-truth count differences (1–7) on \second.} 
    \label{fig:uncertainty-diff} 
\end{figure}

\noindent \textbf{Does model uncertainty reliably predict answer correctness?}
Table \ref{tab:uncertainty} shows that Qwen-7B achieves the most reliable uncertainty scores, despite having the lowest accuracy. This indicates that when this model is highly uncertain, its answer is more likely to be incorrect.
%In other words, when Qwen-7B is highly uncertain about an answer, that answer is more likely to be incorrect.
Qwen-32B has a slightly lower AUROC than Qwen-7B: despite its larger parameter size, this model is more prone to remaining highly confident on \benchmark\  even when it makes mistakes. %, which reduces the separability between correct and incorrect answers.
Finally, GPT-4o mini exhibits the lowest performance, indicating that its uncertainty scores are less reliable for error detection in this setting. 

Figure \ref{fig:uncertainty-diff} shows how AUROC varies for perplexity across ground truth differences in \second, excluding any differences with fewer than 20 samples to promote AUROC stability. For Qwen-7B and GPT-4o mini, we observe a generally increasing trend: when the count difference is small ($diff=1$), %the task is borderline, and 
uncertainty cannot reliably separate right from wrong; as the difference increases ($diff \geq 4$), the comparison becomes clearer, and uncertainty aligns better with errors, so incorrect answers tend to receive higher uncertainty scores. However, this trend does not hold for Qwen-32B: the AUROC at $diff=7$ is even lower than at $diff=1$, despite Qwen-32B reaching a relatively high accuracy of 0.78 for $diff=7$. We hypothesize that in these easier cases, Qwen-32B often produces low uncertainty (high confidence) regardless of correctness, and that this confidence also extends to incorrect examples, reducing the separability between correct and incorrect outputs.

% \noindent \textbf{How does attention analysis explain failures?}
% From initial observations on attention maps of the best-performing model, it generally attends to the correct regions even when producing incorrect answers, consistent with previous findings \cite{zhang2025mllms}. We identify two failure modes illustrated through attention visualizations: first, a \textit{visual ambiguity}/\textit{perception error}, where the model attends to the relevant components but still answers incorrectly; and second, \textit{unnatural errors}, where the model attends to irrelevant regions and produces an incorrect answer. For details, see~\Cref{sec: attention_maps}.

\begin{figure*}[t!] 
    \centering 
    \includegraphics[clip, trim={0 0.2cm 0 0}, width=0.85\linewidth]{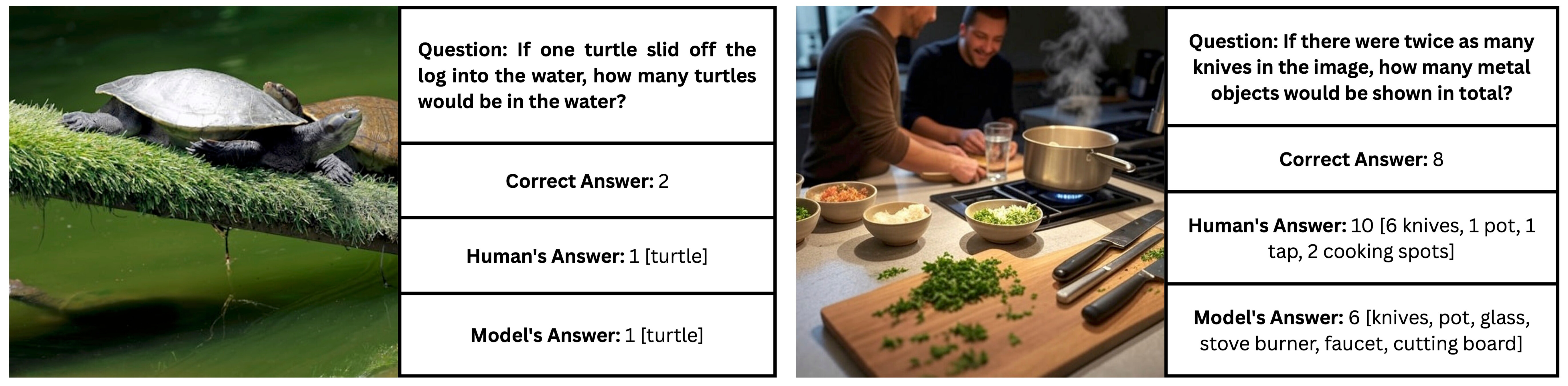} 
    \caption{Examples of errors made by humans and the model. \textbf{Left:} Both the model and the human made a perception error. \textbf{Right: } The human made a semantic ambiguity error, while the model made an unnatural error. } 
    \label{fig:orbit-error} 
\end{figure*}

\noindent \textbf{Do humans and VLMs make similar mistakes?}
% \subsection{Error Analysis}
% We aimed to understand why humans and models make mistakes and whether our model makes errors similar to those made by human annotators. 
% To investigate this, 
We select 25 questions for which both the best VLM (Qwen2.5-VL-32B) and at least one human judge gave an incorrect answer relative to the ground truth. We defined four categories of errors: 1) \textit{Semantic ambiguity} errors caused by unclear semantic constraints in the question, e.g., should an object that is half wood and half metal be counted as ``metal''? 2) \textit{Visual ambiguity} errors caused by ambiguous visual information in the image, e.g., should we count a chair if only part of its legs is visible and not the whole object? 3) \textit{Perception errors} - honest mistakes due to overlooking or misperceiving information, e.g., not noticing a small object or miscounting the number of items. 4) \textit{Unnatural errors} that are highly unlikely for a human to make, e.g., counting a glass as a metal object or counting an object that clearly does not exist. We assign each incorrect answer from the model and the human judge to one or more of these four categories. We distill three key observations (full results in \cref{tab:error_analyzing}).
% The results of this analysis are shown in Table \cref{tab:error_analyzing}.
% We observe the following (full results in \cref{tab:error_analyzing}): a) 

\begin{figure}[tb]
  \centering
  % % --- First Subfigure ---
  \begin{subfigure}[t]{0.85\linewidth}
    \centering
    \includegraphics[width=0.85\linewidth]{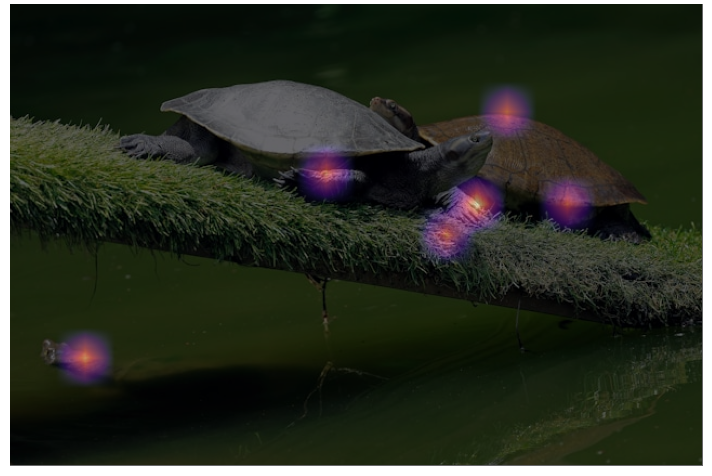}
    \caption{How many reptiles are visible in this image? \textbf{G}: 3, \textbf{M}: 2}
    \label{fig:short-evo-a}
  \end{subfigure}
  \hfill
  \hfill
  % --- Second Subfigure ---
  \begin{subfigure}[t]{0.85\linewidth}
    \centering
    \includegraphics[width=0.85\linewidth]{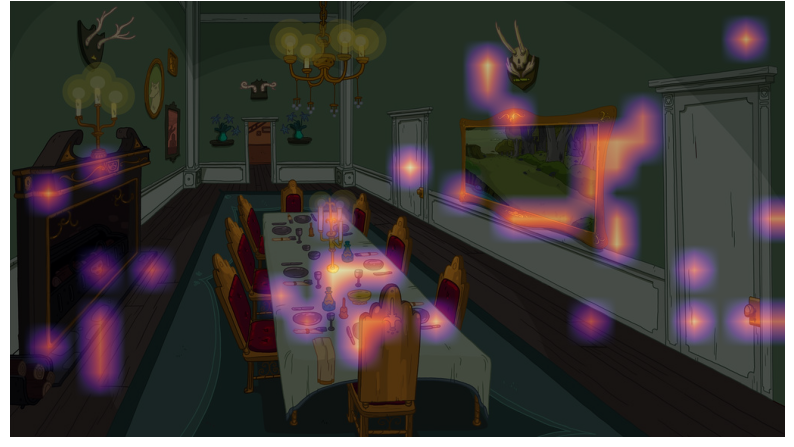}
    \caption{If all the paintings were replaced with mirrors, how many reflective surfaces would be visible? \textbf{G}: 4, \textbf{M}: 5}
    \label{fig:short-img-b}
  \end{subfigure}
  % \hfill
  % % --- Second Subfigure ---
  % \begin{subfigure}[t]{0.32\linewidth}
  %   \centering
  %   \includegraphics[width=\linewidth]{img/CNT_qwen2-5-vl-32b_attention_overlay_REAL_image08_Q2_layer54.png}
  %   \caption{How many objects are present that act as support? \textbf{G}: 1, \textbf{M}: 1}
  %   \label{fig:short-evo-b}
  % \end{subfigure}
  % \hfill
  % % --- Third Subfigure ---
  % \begin{subfigure}[t]{0.32\linewidth}
  %   \centering
  %   \includegraphics[width=\linewidth]{img/CNT_qwen2-5-vl-32b_attention_overlay_REAL_image08_Q3_layer54.png}
  %   \caption{If one turtle slid off the log into the water, how many turtles would be in the water? \textbf{G}: 2, \textbf{M}: 1}
  %   \label{fig:short-evo-c}
  % \end{subfigure}
  
  % --- Main Figure Caption and Label ---
  \caption{Qwen2.5-VL (32B) attention maps (layer 54 and 64, respectively) for examples from \first. Showcasing the two failure modes of visual ambiguity/perception errors in (a) and unnatural errors in (b). \textbf{G} and \textbf{M} represent the ground truth and model predictions, respectively.}
  \label{fig:short_CNT_evo}
\end{figure}

First, \textit{humans do not make unnatural mistakes (0/25), whereas the model makes many of them (9/25)}. This is the most significant difference between humans and the model, which may stem from hallucination/illusion-driven errors in models, as evidenced by diagnostic suites~\cite{guan2024hallusionbench}. An example is shown in \cref{fig:orbit-error} (right), where the human counted all objects that could be considered metal, while the model counted items like the glass and the cutting board, which are clearly not metal. 
We further investigate the nature of model errors by examining attention maps of the best-performing model, which provide complementary evidence for the typical model failure modes (unnatural errors and perceptual ambiguity errors). 
Namely, the Qwen model generally attends to the correct regions even when producing incorrect answers, consistent with previous findings \cite{zhang2025mllms}. In \cref{fig:short-evo-a}, the model's attention broadly covers the turtles on the log but undercounts them, consistent with our earlier finding that both the model and human judges overlooked the turtle partially submerged in the water. In \cref{fig:short-img-b}, the model overcounts reflective surfaces, with attention hotspots scattered across regions that bear no correspondence to actual or hypothetical mirrors, pointing to object hallucination and consistent with an unnatural error. Finally, attention visualizations on \second\ (\cref{fig:short_CMP}) show that attention spans both images in the pair, validating the multi-image setup.
%We provide further insights into how attention patterns shift with absolute-counting and multi-image comparison tasks in \Cref{sec: attention_maps}.

Second, \textit{for humans, the majority of mistakes (16/25) come from semantic ambiguity}, not knowing whether something should be counted or how to interpret a specific condition. This finding is consistent with our analysis of image types, which concluded that photographic images may be most challenging due to remaining ambiguity in the questions or in the visual content.
% Consistent with this, our extended error analysis to understand whether humans and models make similar mistakes shows that most human errors stem from semantic ambiguity in the questions or in the visual content (see Figure \cref{fig:orbit-error} and Table \cref{tab:error_analyzing}).

Third, \textit{the second most common source of human error is perception errors} (9/25), often due to not examining the image carefully enough and missing objects. Perception errors are similarly frequent in humans and the model (9 and 8, respectively). An example is shown in \cref{fig:orbit-error} (left), where both the human and the model made the same mistake: they missed the turtle inside the water.

% hallucination/illusion-driven errors (e.g., false positives in clutter) in light of evidence from controlled diagnostic suites (Guan et al., 2024)

\begin{table}[t!]
\caption{Micro-accuracy results for frontier and reasoning models on OPTICS. GPT-o3 is a reasoning model.}
\label{tab:frontier_model_results}
\centering
% \scriptsize
\setlength{\tabcolsep}{1pt}
\begin{tabular}{l|r|r}
\toprule 
\textbf{Model} & 
 \textbf{OPTICS-CNT ($\uparrow$)}
& \textbf{OPTICS-CMP ($\uparrow$)} \\
\midrule
Qwen3.5 (9B) & 44.17 & 58.39 \\
% Qwen3.5 - thinking (9B) (67\% data)  & $\checkmark$  &  & 81.28 \\
% & 
Gemini-2.5-flash  & 50.19 & 61.50  \\
% & 
GPT-o3 & \textbf{52.41} & \textbf{82.78}  \\ \hline
% \midrule
\end{tabular}
% }
% \vspace{-2em}

\end{table}

\noindent \textbf{How do frontier and reasoning models perform on OPTICS?}
In addition to the baseline models, we evaluated three newer models on our benchmark: Qwen3.5-9B \cite{qwen3.5}, Gemini 2.5 Flash \cite{comanici2025gemini}, and OpenAI o3 \cite{openai_o3_o4mini_system_card}. The results are reported in \cref{tab:frontier_model_results}. In general, compared to the results in tables \ref{tab:model_results} and \ref{tab:model_results_compare}, these frontier models achieve higher precision in both \first\ and \second. In particular, OpenAI's o3 achieves the strongest performance, reaching 52.41 on \first\, compared to the previous best result of 39.91. It also reaches 82.78 on \second\, improving on the previous best result of 68.33. Overall, this corresponds to a notable improvement of approximately 12-14\% across the two datasets. However, its performance on \first \ remains around 20\% below that of humans. Additional error analysis shows that unnatural errors are uncommon for GPT-o3 compared to other models. In summary, these results suggest that these models achieve substantially higher accuracies and make fewer unnatural errors, but still fail to achieve human-level performance.

\section{Conclusions and Future Directions}% (0.5)}

This paper developed a commonsense evaluation framework for evaluating object property reasoning in images, covering three distinct image types (photographic, animated, and AI-generated), three reasoning levels (direct recognition, property inference, counterfactual), and four property types (physical, taxonomic, functional, relational). We instantiated this framework on two VQA benchmarks: \first, with over 1K counting questions, and \second, with over 2K comparison questions.
%split uniformly across the evaluation dimensions.
% We introduced \benchmark, a VQA benchmark focused on reasoning about object properties through simple count-based questions. \benchmark has 360 images (real, animated, and AI-generated) and 1,080 questions across three reasoning levels (direct recognition, property inference, counterfactual) and four property types (physical, taxonomic, functional, relational). 
Experiments with 12 state-of-the-art VLMs revealed low counting accuracy, with most models performing far below human performance and the best achieving only 40\% accuracy, which increased to 52\% when including GPT-o3.
VLMs struggle particularly with photographic images, counterfactual reasoning about physical and functional properties, and infrequent counts greater than five.
Humans also struggle with photographic images and physical properties, but excel at counterfactual questions and have no frequency effects on counts.
Although the samples of \second\ are derived from \first, we note that it evaluates a different reasoning capability over counts, yielding new insights: whether models perform approximate comparisons for larger count differences, how model uncertainty can distinguish correct from incorrect answers, and where model attention is focused.
Interestingly, when asked to compare images based on their counts, most VLMs performed better for higher differences, but their performance on very large differences remained relatively modest. Analyzing the errors made by humans and VLMs closely revealed that, although both make a sizeable proportion of perceptual mistakes, VLMs are dominated by unnatural errors, whereas humans struggle with semantic ambiguity. 
% This indicates that VLMs experience limitations on a combination of perception and reasoning abilities and unexpected hallucinations.
% \todo{add findings from orbit-compare}
% Results show particular difficulty with physical and relational properties, and while models struggle with direct recognition, they do better with abstract reasoning. 
% Real-world images posed greater challenges than animated or AI-generated ones. 
% Proof-of-concept experiments with a spatially adapted VLM show the promise of property-specific fine-tuning.

In summary, unlike previous benchmarks, %(TallyQA, HowmanyQA, TDIUC, AOKVQA, VCR, NLVR2), 
OPTICS jointly controls the three axes of object properties, reasoning complexity, and image types. This leads to novel insights: (1) \textit{Perception bottleneck}: reasoning complexity ordering inverts on top-tier VLMs, exposing perception, not reasoning, as the bottleneck. (2)\textit{ Realism gap}: the ordering of task difficulty across image types holds for both humans and VLMs and is `animated < AI-generated < photographic'. (3) \textit{Error tendencies}: while VLMs make unnatural mistakes, humans struggle with semantic ambiguity.

We consider three directions for future work to improve on the work presented here. %should be addressed in.
First, generative AI still suffers from limited accuracy and diversity in autonomous question generation, causing \textit{scaling up object reasoning benchmarks to require substantial human effort}. % to develop bench. 
% despite the careful framework design and data collection procedure, it remains challenging to generate high-quality benchmarking data at scale without substantial human effort. 
% Generative AI advances and existing rich annotations speed up the procedure to some extent, however, the resulting dataset still suffers from limited size and diversity. 
Our attempt to generate a scalable version based on richly annotated data (\cref{sec:vg_exp}) showed that automatically creating such benchmarks remains challenging.
Novel approaches,
% Methods, 
such as metamorphic testing~\cite{yuan2021perception}, are necessary to grow \first\ and \second~systematically in size and coverage of additional object properties, question types beyond counting and comparison, and open-ended answers. The quality of VLM-generated questions can be improved by incorporating self-correction mechanisms~\cite{liao2025can}, human visual prompting~\cite {cai2024vip}, or recent automated scoring metrics~\cite{aghazadeh2025cap}. Thus, while our primary focus is on providing high-quality evaluation and detailed analysis, we see scaling up as an important direction for future work.

% Future work comprises expanding the benchmark with more images and question diversity across the four property dimensions and including binary and open-ended question formats to better evaluate models' conceptual understanding, revealing any inductive biases that might persist. 
Second, since annotators are free to design their own questions, they may \textit{introduce semantic ambiguity}, leading to human disagreements (e.g., whether zebras are considered black animals). Although we addressed ambiguity through an explicit quality-assurance step that analyzed sources of disagreement and translated them into more precise guidelines, some disagreements remain. Future work should investigate how to develop more generalizable guidelines for object abstraction and reasoning to further reduce semantic ambiguity.
% For instance, when asked to identify an object \textit{with handles}, a bike, a tennis racket, and a backpack are considered, but a backpack could also be understood as an object \textit{having straps}. 

Finally, while \first\ and \second\ point to fundamental limitations of popular and diverse
% provides complementary insights to prior benchmarks into the common limitations of
VLMs, %regardless of their architecture, size, or training, 
\textit{the set of baselines is inherently incomplete}. Future work should study other specialized reasoners beyond SpatialVLM, including a reference expression counting method that addresses compositional attribute filters and counting \cite{dai2024referring}, mixtures of experts~\cite{lin2024moe}, derivation of object schemas that support generalization \cite{hsu2024makes}, reinforcement learning methods that separate perception from reasoning~\cite{jiang2025videop2r}, and methods for populating situational scene graphs for preset schemas \cite{sugandhika2025situational}. %, and reasoning over ~\citep{},  

\section*{Acknowledgements}

This paper is based on Abhishek Kolari's MSc AI project at Vrije Universiteit Amsterdam.
MK and FI were funded by the AiNed project \textit{Human-Centric AI Agents with Common Sense} supported by the Dutch Organisation for Scientific Research (NWO), under grant no. NGF.1607.22.044.
FdH was funded by the Hybrid Intelligence Centre, a 10-year programme supported by the Dutch Ministry of Education, Culture and Science through NWO (grant no. 024.004.022). The evaluations were performed on the DAS-6 cluster~\citep{bal2016medium}.

%\printcredits
\bibliographystyle{cas-model2-names}
\bibliography{ijcai22}

\appendix
\clearpage

\appendix
\section{Appendix: Technical Details}

\begin{table*}[hb!]
\caption{Prompt template used to extract output in a specific format from the model for non-instruction-tuned and instruction-tuned models. \textit{\{IMG\}} and \textit{\{question\}} are the placeholders for the image and question, respectively.}
\label{tab:output_prompts}
\centering
% Adjust column separation. Default is usually 1pt. Try a larger value.
\setlength{\tabcolsep}{5pt}
% \resizebox{\textwidth}{!}{%
% \makebox[\linewidth][c]{%
\begin{tabular}{l|p{0.7\linewidth}}
\toprule
\textbf{Category} & \textbf{Prompt Template}\\
\midrule
Non-Instruction-tuned models & \textit{\{IMG\}}\newline Question: \textit{\{question\}} Provide only the total number. Answer:\\
\midrule
Instruction-tuned models & \textit{\{IMG\}}\newline\textit{\{question\}} Your response MUST be in the following format and nothing else: $<$NUMBER$>$ [$<$OBJECT1$>$, $<$OBJECT2$>$, $<$OBJECT3$>$, ...]\\
\bottomrule
\end{tabular}
% } %
\end{table*}

\begin{table*}[hb!]
\caption{Prompt examples used to generate AI images across different domains, along with corresponding tags (shown in brackets) used in constructing the MacGyver dataset. Prompts with the \textbf{Create Image} tool are generated using GPT-4o, the rest with Grok 3.}
\label{tab:aigen_prompts}
\centering
\small
% Adjust column separation. Default is usually 1pt. Try a larger value.
\setlength{\tabcolsep}{5pt}
% \resizebox{0.9\textwidth}{!}{%
% \makebox[\linewidth][c]{%
\begin{tabular}{p{0.08\linewidth}|p{0.85\linewidth}}
\toprule
\textbf{Image Domain} & \textbf{Prompt Example}\\
\midrule %\cmidrule{5-8} \cmidrule{9-11}
Campsite settings (Outdoors) & \textbf{Create image} Generate a photorealistic dusk campsite scene with 5 tents, a few of them illuminated from within, a controlled campfire with 3 people sitting on logs around it and a dog sleeping nearby, hiking backpacks leaning against a nearby tree, a lantern hanging from a branch, and stars beginning to appear in the darkening sky. Include a mountain silhouette in the background. Strictly keep the count of total objects in the image to a maximum of 10.\\
\midrule
Gym (Indoors) & \textbf{Create image} Generate a photorealistic image of a gym scene with sports equipment. Let the gym contain 3 treadmills, 2 elliptical machines, a few kettlebells on the floor, 2 resistance bands hanging on the wall, and a bench press rack. Let it be a realistic setting with 2 humans working out, one person on the treadmill and another person on the elliptical machine. The gym can have a clock and TV hanging on the wall somewhere showing a workout video. Include large windows with natural light and a small plant in the corner. Strictly keep count of total objects in the image to a maximum of 10.\\
\midrule
Beach cleanup (Outdoors) & Generate a photorealistic image of a beach cleanup scene with 4 volunteers in bright t-shirts collecting trash at sunset in the backdrop. Let the trash include these objects: 5 glass bottles, 6 trash bags (one in each hand of a volunteer and 2 placed near the ground), a collection bin, reusable gloves, and garbage pickers. Let it be a realistic setting, having the ocean with gentle waves in the background. Strictly keep the count of total objects in the image to a maximum of 10.\\
\midrule
Garage (Neutral) & \textbf{Create image} Create a photorealistic image of a vehicle garage scene during a vehicle maintenance session. Show 3 cars, one car on a hydraulic lift with its hood open, 2-3 essential tools spread on a workbench, an oil change in progress, spare parts in organised containers, and a service manual. Let the setting be realistic with ceiling lights for the garage, let the car being serviced be in good condition, and let the other two cars in the background be scrap and old, which need to be rebuilt with loose tires leaning next to them. Let one person work on the car that is being serviced and the other person in the backdrop next to the other 2 cars. Strictly keep count of total objects in the image to a maximum of 10.\\
\midrule
Kitchen (Indoors) & \textbf{Create image} Generate an image of a kitchen countertop during the preparation of a complex dish. Show fresh ingredients arranged in 3 small ceramic bowls, a wooden cutting board with chopped herbs, a professional knife set, a simmering pot on the stove, a recipe book, a glass filled with water, and 2 people collaborating on the cooking. Include details like steam, a kitchen sink next to the stove, and ambient kitchen lighting. Let it be a realistic setting. Strictly keep the count of total objects in the image to a maximum of 10.\\
\bottomrule
\end{tabular}
% } %
\end{table*}

\subsection{Prompt Strategies}

\subsubsection{Question Generation}
\label{ssec: Appendix_prompt_question}

We pass a standard prompt template to any of the three MLLMs: GPT-4o, Claude 3.7 Sonnet, and Gemini 2.0 Flash to generate questions, which are then used to inspire humans to generate new questions manually. The template is as follows:

\textit{\{IMG\} Generate three questions per prompt for this image with answers for each prompt. Prompt 1 deals with physical/taxonomic properties; prompt 2 deals with functional/relational properties; and prompt 3 deals with counterfactual reasoning. All need to be counting questions; be creative. You do not need to stick to the same questions.}

% shown in Figure \cref{fig:question_prompt}.

% \begin{figure}[ht]
%     \centering
%     \textit{\{IMG\} Generate three questions per prompt for this image with answers for each prompt. Prompt 1 deals with physical/taxonomic properties; prompt 2 deals with functional/relational properties; and prompt 3 deals with counterfactual reasoning. All need to be counting questions; be creative. You do not need to stick to the same questions.}
%     \caption{Standard prompt template for question generation, where \textit{\{IMG\}} is the placeholder for the image. The instruction is to generate a variety of questions with answers in the same format as \benchmark, with the 'prompt' in the template referring to the three reasoning levels.}
%     \label{fig:question_prompt}
% \end{figure}
% \vspace{-2em}

\subsubsection{Output Formatting}
\label{ssec: Appendix_prompt_output}

\Cref{tab:output_prompts} displays the prompt templates used on instruction-tuned and non-instruction-tuned models to extract outputs in a specific format. Instruction-based models are also asked for a list of objects detected and considered for the count, which  
% but this information is not included in the paper as it 
is currently not directly evaluated. % part of the current evaluation setup but likely for future work. 

\subsubsection{AI-Generated images}
\label{ssec: Appendix_prompt_aigen}

\Cref{tab:aigen_prompts} shows the prompt examples used while generating AI-based images for different image domains.

\begin{table}[!t]
\caption{The 15 domains (i.e., locations and activities) taken from the MacGyver dataset~\cite{tian2025macgyverlargelanguagemodels} (marked by an asterisk), extended with 11 of our own, for a total of 26 unique image domains. They are broadly divided into Indoors, Neutral, and Outdoors.}
\label{tab:image_theme_tags}
\small
\centering
% Adjust column separation. Default is usually 1pt. Try a larger value.
\setlength{\tabcolsep}{5pt}
\makebox[\linewidth][c]{%
\begin{tabular}{c|c|c}
\toprule
\textbf{Indoors} & \textbf{Neutral} & \textbf{Outdoors}\\
\midrule
bedroom* & tools & beach cleanup*\\
home setup & tech & camping*\\
kitchen* & garage* & construction*\\
library* & classroom* & gardening*\\
laboratory & & biking\\
meeting room* & & market\\
gym* & & zoo*\\
salon & & urban\\
wardrobe* & & park*\\
& & picnic\\
& & farm*\\
& & driveway\\
& & bustop\\
\bottomrule
\end{tabular}
}
\end{table}

\subsection{Image Domains}
\Cref{tab:image_theme_tags} shows the list of 26 unique image domains used for the curation of \benchmark. 

\subsection{Additional Benchmark Examples}
\label{sec: benchmark_examples}

% \subsubsection{Image Types}

\begin{figure}[th!] % [h!] is a placement specifier: here, top, bottom, page of its own
    \centering % Centers the image horizontally on the page
    \includegraphics[width=\linewidth]{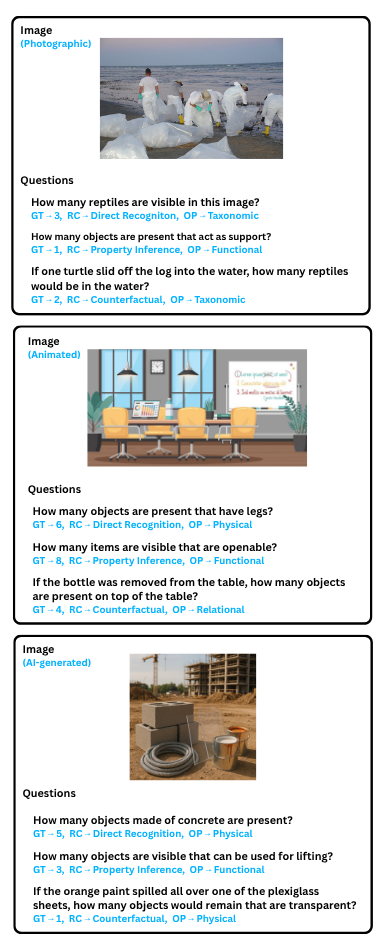} % Replace with your image file name
    \caption{Examples from \first, for three image types: Photographic (top), Animated (middle), and AI-generated (bottom). Text highlighted in \textcolor{cb}{blue} is information stored in the benchmark and not passed to the model during evaluation. \textcolor{cb}{GT} stands for ground truth, \textcolor{cb}{RC} for reasoning complexity, and \textcolor{cb}{OP} for object properties.} % Caption for the image
    \label{fig:benchmark_examples} % Label for cross-referencing (e.g., "as seen in Figure \cref{fig:simple_image}")
\end{figure}

% \section{Category-Specific Examples}
% \label{sec: cat_examples}

\subsubsection{\first}
% \label{ssec: objprop_examples}

\Cref{fig:benchmark_examples} shows additional examples from \first\ for three image types: Photographic, Animated, and AI-generated.
\Cref{tab:objprop_questions} shows examples of questions present per object property dimension.

\begin{table*}[!ht]
\caption{Question examples for each of the four object property dimensions.}
\label{tab:objprop_questions}
\small
\centering
% Adjust column separation. Default is usually 1pt. Try a larger value.
\setlength{\tabcolsep}{5pt}
\makebox[\linewidth][c]{%
\begin{tabular}{c|p{0.7\linewidth}}
\toprule
\multicolumn{1}{c|}{\textbf{Dimensions}} & \textbf{Question Examples}\\
\midrule
\multirow{3}{*}{Physical} & How many objects made of wood are present?\\ & How many objects are present that are transparent?\\ & How many objects in the background are present that have legs?\\
\midrule
\multirow{3}{*}{Taxonomic} & How many mammals are visible in the image?\\ & How many furniture items are present in the room?\\ & How many tools are visible in the image?\\
\midrule
\multirow{3}{*}{Functional} & How many objects with the primary purpose of illumination can be seen?\\ & Count the number of breakable items?\\ & Count the number of items that are battery powered?\\
\midrule
\multirow{2}{*}{Relational} & How many objects are visible that are attached to the wall or ceiling?\\ & How many reptilian couples, at maximum, are present?\\
\bottomrule
\end{tabular}
}
\end{table*}

\begin{figure}[ht!] % [h!] is a placement specifier: here, top, bottom, page of its own
    \centering 
    \includegraphics[width=\linewidth]{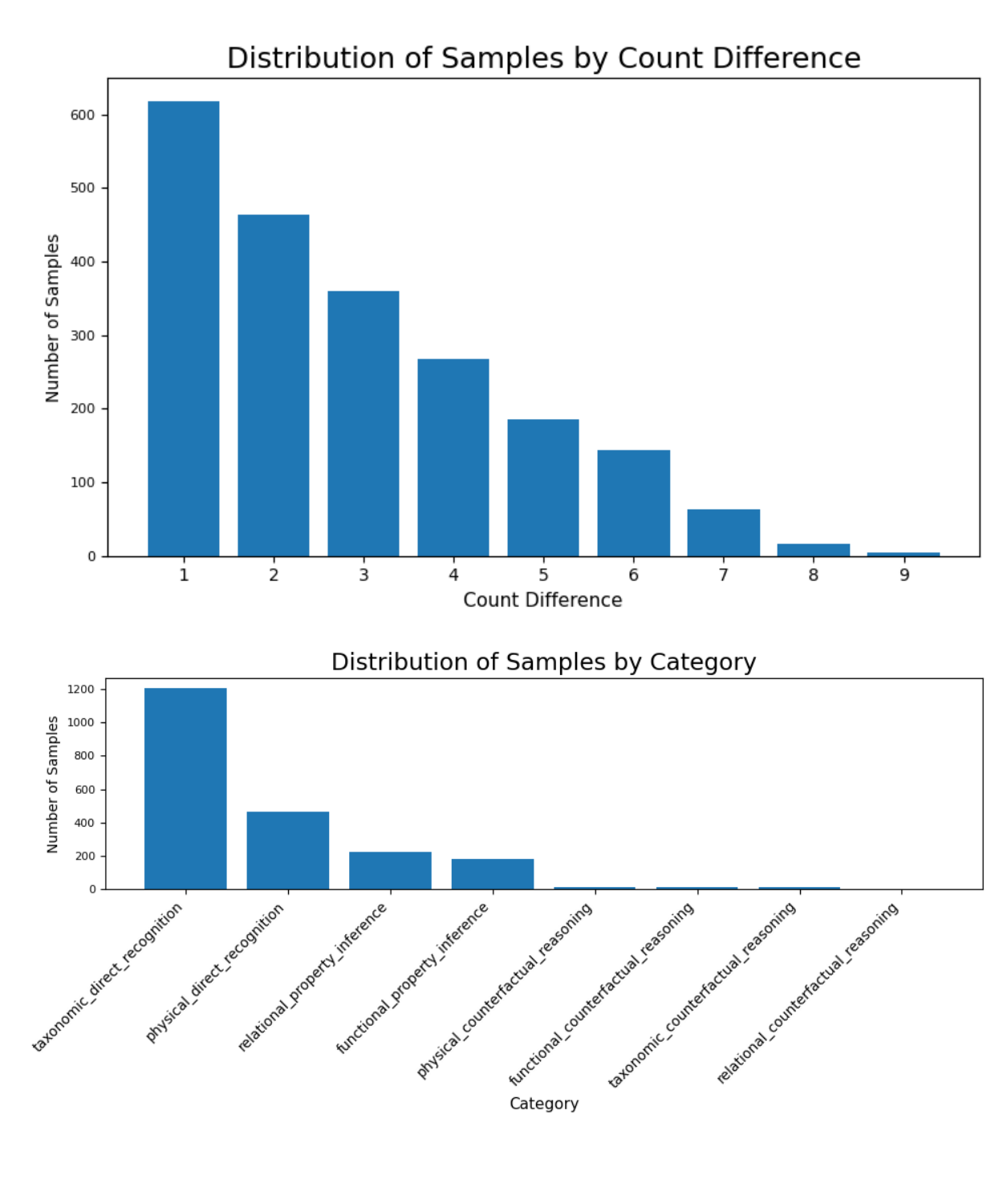} 
    \caption{Sample distribution of \second: count difference (top) and distribution by category (bottom).} % Caption for the image
    \label{fig:orbit_compare_statistic} 
\end{figure}

\begin{figure}[ht!] % [h!] is a placement specifier: here, top, bottom, page of its own
    \centering % Centers the image horizontally on the page
    \includegraphics[width=\linewidth]{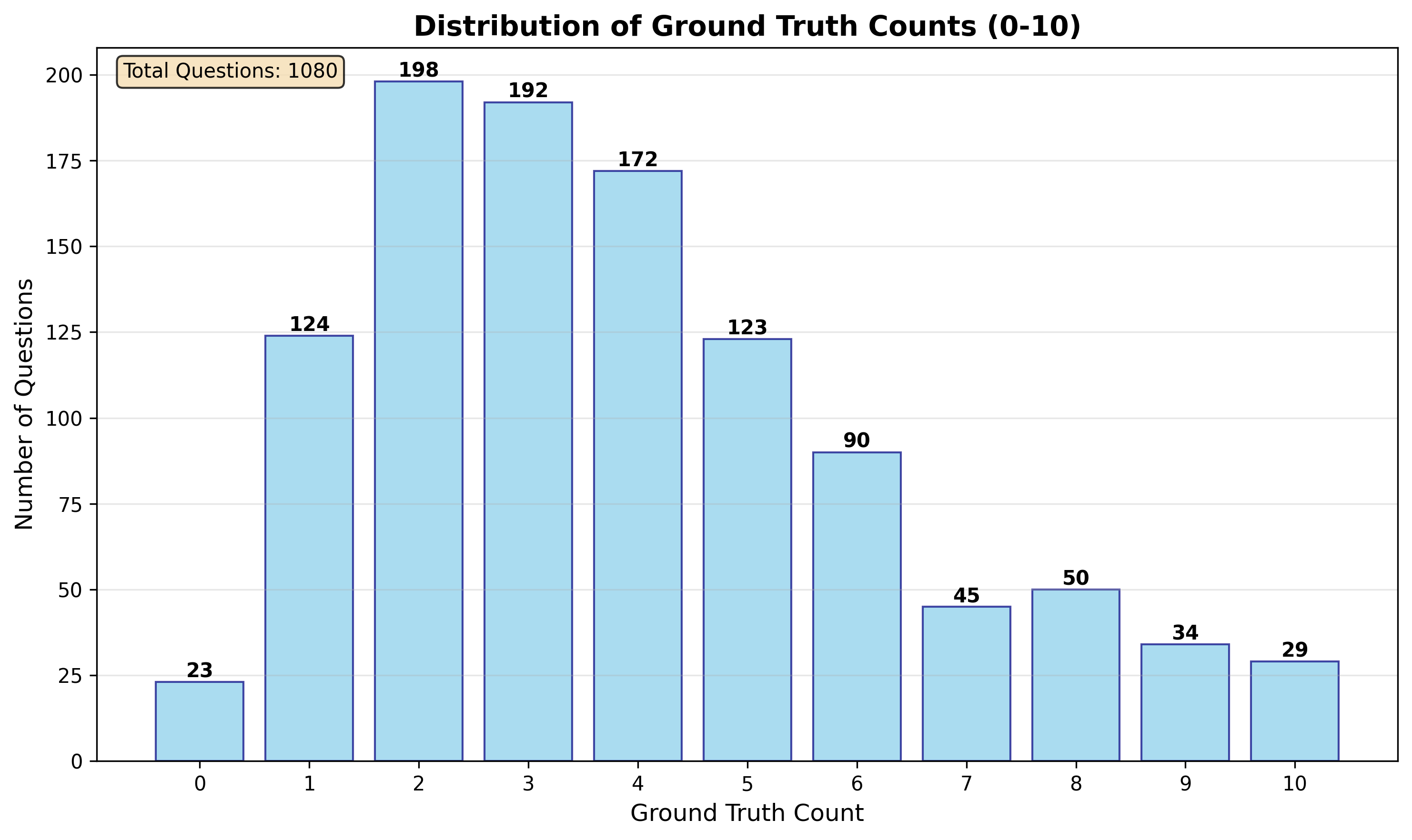} % Replace with your image file name
    \caption{Ground truth distribution for all counts (0 to 10).} % Caption for the image
    \label{fig:gt_dist} % Label for cross-referencing (e.g., "as seen in Figure \cref{fig:simple_image}")
\end{figure}

% \vspace{-2em}
\subsubsection{\second}
\label{ssec: Orbit_compare_example}
\Cref{fig:orbit_compare_example} shows an example from \second\ where we merge two images from \first\ into a single image. \Cref{fig:orbit_compare_statistic} visualizes the sample distribution across count-difference levels and question categories.

\subsection{Ground Truth Distribution}
\label{sec: GT_distribution}

For the total number of 1,080 questions in \first, the ground truth distribution for all the counts ranging from 0 to 10 is shown in \cref{fig:gt_dist}.

\section{Appendix: Complementary Results}

\subsection{Cumulative Accuracies with Count Tolerance}

\begin{figure}[!t] % [h!] is a placement specifier: here, top, bottom, page of its own
    \centering % Centers the image horizontally on the page
    \includegraphics[width=\linewidth]{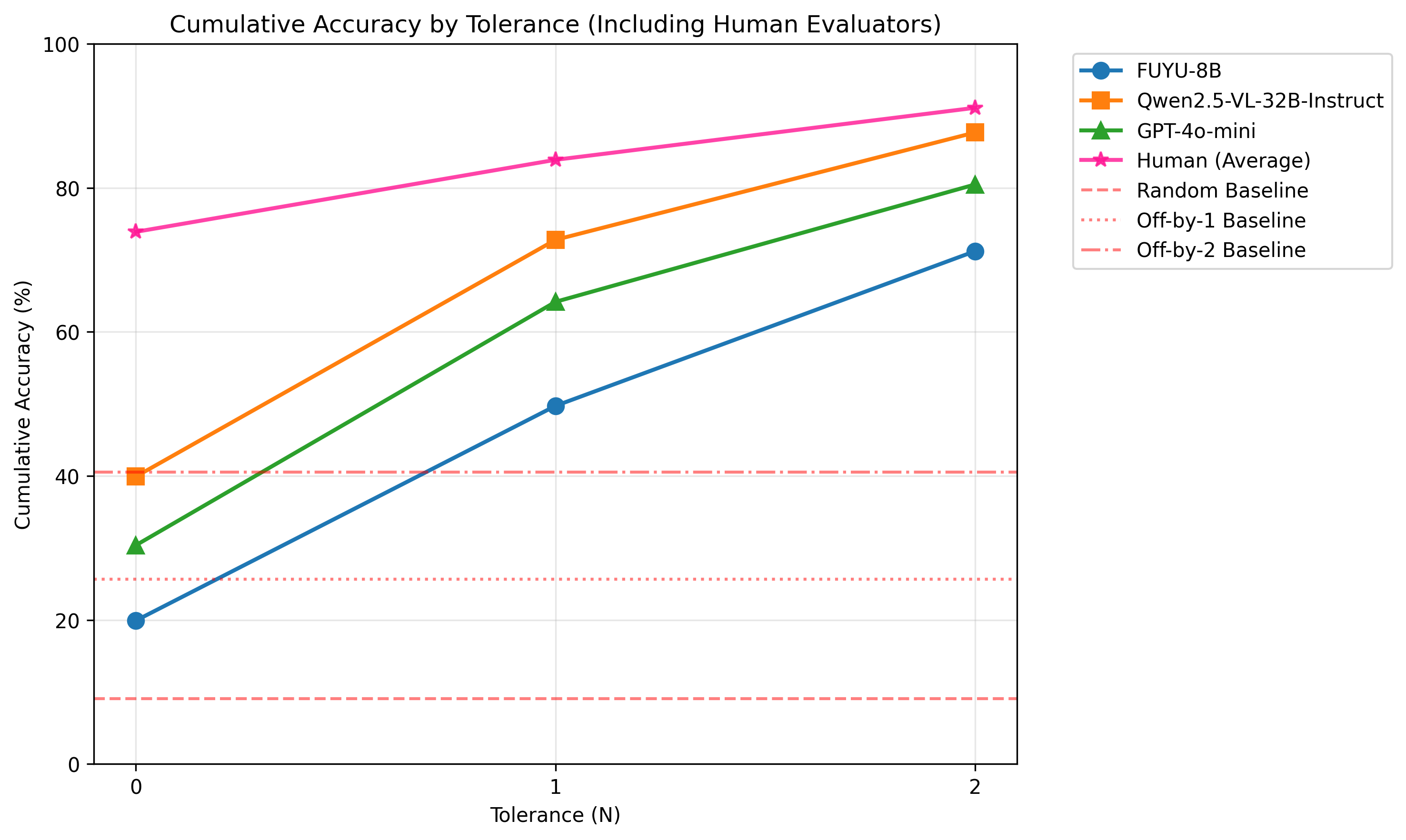} % Replace with your image file name
    \caption{Cumulative \first accuracies for tolerance levels 0, 1, and 2. The top-performing models from the following categories are shown: open-source non-instruction VLM (Fuyu-8B), open-source instruction VLM (Qwen2.5-VL-32B-Instruct), closed-source VLM (GPT-4o mini), the human evaluators' average and the random chance baselines for tolerance levels 0 (9.09\%), 1 (25.62\%), and 2 (40.50\%).} % Caption for the image
    \label{fig:cumulative_accuracies} % Label for cross-referencing (e.g., "as seen in Figure \cref{fig:simple_image}")
\end{figure}

The soft off-by-1 accuracy metric, which has a baseline of 25.62\%, shows relatively larger improvements (a 47.16\% gap over the best model) among the larger instruction-tuned models. Even though the models' predictions are often nearly correct, this metric highlights the difficulty in precise numeric predictions by VLMs. The cumulative accuracy for tolerance values 0, 1, and 2 is shown in \cref{fig:cumulative_accuracies} for the models that perform the best in their categories and the average of the human evaluators.

\subsection{Inductive Bias on \first}
\label{sec:Inductive_bias}

\begin{figure}[!t] % [h!] is a placement specifier: here, top, bottom, page of its own
    \centering % Centers the image horizontally on the page
    \includegraphics[width=0.85\linewidth]{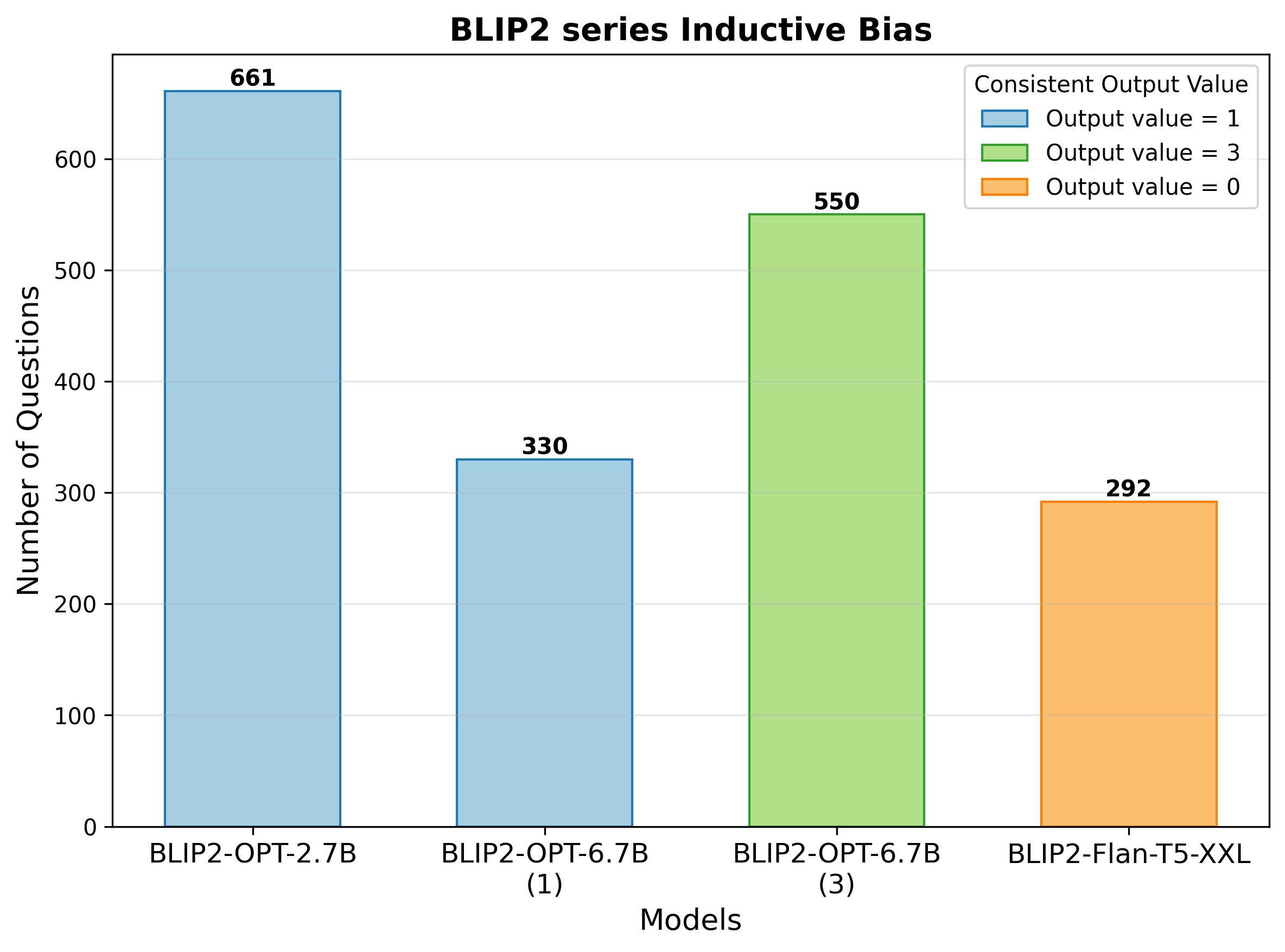} % Replace with your image file name
    \caption{BLIP2 models' inductive bias on \first, consistently defaulting to certain output values.} % Caption for the image
    \label{fig:inductive_bias} % Label for cross-referencing (e.g., "as seen in Figure \cref{fig:simple_image}")
\end{figure}

% For instance, the BLIP2 family with its OPT 2.7B variant consistently (frequency distribution included in the Appendix) predicted the value "1", the OPT 6.7B predominantly predicted either "1" or "3", and the Flan T5-xxL predicted "0" for most questions 
% \vspace{-150pt}
The BLIP2 model series consistently defaults to specific output values. The OPT 2.7B variant prefers the count 1, the OPT 6.7B variant produces either 1 or 3 as the output, and the Flan T5 xxL gives an output of 0 or no value or an answer not abiding by the required format (numeric value), resulting in a final output of 0 set according to our answer extraction method. The frequency distribution of these outputs is shown in \cref{fig:inductive_bias}.
We tried adjusting prompts and decoding parameters, such as the maximum number of new tokens, but none of them helped mitigate the issue, highlighting the inductive biases in models \cite{wang2023theoretical, kraaijveld2025columbus}.

\begin{table}[!t]
\caption{Analysis of errors made by humans and Qwen2.5 (32B).}
\label{tab:error_analyzing}
\centering
\small
% \resizebox{\linewidth}{!}{
\begin{tabular}{l|r|r|r|r}
\toprule
\textbf{} &
\parbox{1.3cm}{\centering \textbf{Semantic\\ambiguity}} &
\parbox{1.3cm}{\centering \textbf{Visual\\ambiguity}} &
\parbox{1.4cm}{\centering \textbf{Perception\\errors}} &
\parbox{1.25cm}{\centering \textbf{Unnatural\\errors}} \\
\midrule
Human         & 16 & 3 & 9 & 0 \\
Qwen2.5 &  8 & 1 & 8 & 9 \\
\bottomrule
\end{tabular}
% }
\end{table}

% Our initial observations on attention maps of the best-performing model indicate that it generally attends to the correct regions even when producing incorrect answers, consistent with previous findings \cite{zhang2025mllms}. We identify two failure modes illustrated through attention visualizations: first, a \textit{visual ambiguity}/\textit{perception error}, where the model attends to the relevant components but still answers incorrectly; and second, \textit{unnatural errors}, where the model attends to irrelevant regions and produces an incorrect answer. 

% In \cref{fig:short-evo-a}, the model's attention broadly covers the turtles on the log but undercounts them, consistent with our earlier finding that both the model and human judges overlooked the turtle partially submerged in the water. In \cref{fig:short-img-b}, the model overcounts reflective surfaces, with attention hotspots scattered across regions that bear no correspondence to actual or hypothetical mirrors, pointing to object hallucination and consistent with an unnatural error. Finally, attention visualizations on \second\ (\cref{fig:short_CMP}) show that attention spans both images in the pair, validating the multi-image setup.

\subsection{Error Analysis of Humans and Qwen}

The results of our analysis comparing human and Qwen errors are provided in \cref{tab:error_analyzing}.

\begin{figure}[!th]
  \centering
  % --- First Subfigure ---
  \begin{subfigure}[t]{\linewidth}
    \centering
    \includegraphics[width=\linewidth]{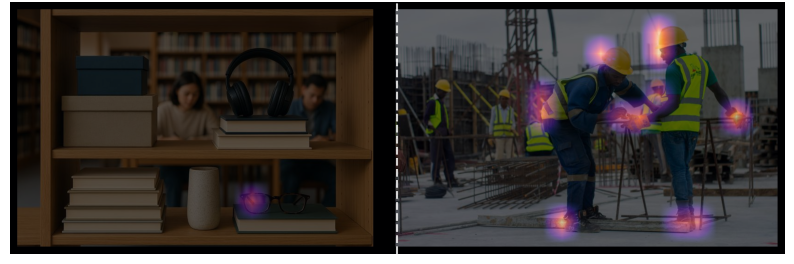}
    \caption{Let A be the answer to "How many objects are present that can be worn on the head?" for the left image. 
Let B be the answer to "How many objects are visible that are designed to protect the head?" for the right image. Which is greater, A or B? \textbf{G}: B, \textbf{M}: B}
    \label{fig:short-cmp-a}
  \end{subfigure}
  \hfill
  % --- Second Subfigure ---
  \begin{subfigure}[t]{\linewidth}
    \centering
    \includegraphics[width=\linewidth]{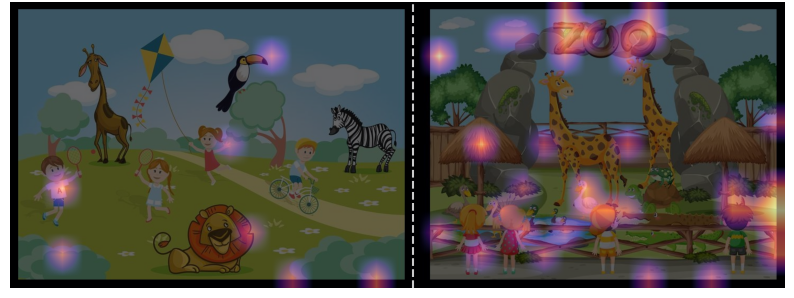}
    \caption{Let A be the answer to "How many animals are present?" for the left image. 
Let B be the answer to "How many animals that are birds are present?" for the right image. Which is greater, A or B? \textbf{G}: B, \textbf{M}: A}
    \label{fig:short-cmp-b}
  \end{subfigure}
  
  % --- Main Figure Caption and Label ---
  \caption{Qwen2.5-VL (32B) attention maps (layer 52) across images from \second. \textbf{G} and \textbf{M} represent the ground truth and model predictions, respectively.}
  \label{fig:short_CMP}
\end{figure}

\subsection{\second\ Attention Maps}

% We report the overall micro accuracy and the macro accuracy, which is the average across all count differences (1-9).
% macro-accuracy 
% across count differences, expecting to see increasing performance as the count difference increases.
Representative attention maps of Qwen2.5-VL-Instruct, our best-performing model, for questions from \second, are shown in \cref{fig:short_CMP}.

\begin{figure}[t!] % [h!] is a placement specifier: here, top, bottom, page of its own
    \centering % Centers the image horizontally on the page
    \includegraphics[width=\linewidth]{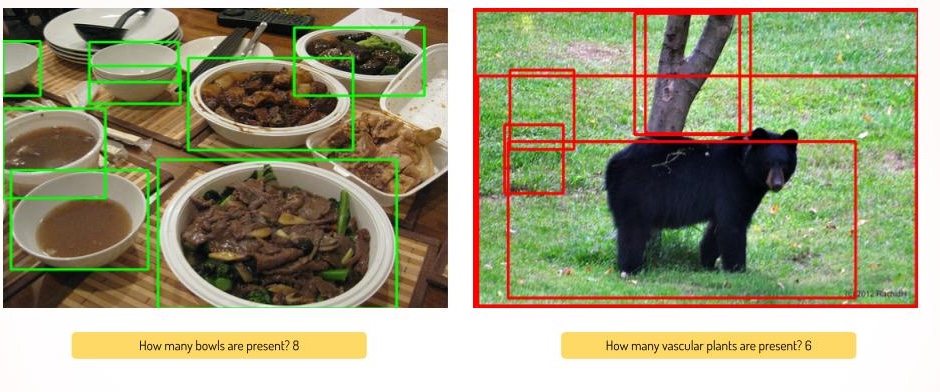} % Replace with your image file name
    \caption{\textcolor{cg}{Left}: A successful example of a question generated using Visual Genome.
    \textcolor{cr}{Right}: An unsuccessful example, where multiple objects are assigned to a single tree and a single patch of grass, leading to an incorrect answer.} % Caption for the image
    \label{fig:orbit-vg} % Label for cross-referencing (e.g., "as seen in Figure \cref{fig:simple_image}")
\end{figure}

\subsection{Visual Genome-based Dataset Curation}
\label{sec:vg_exp}

We explore whether densely annotated datasets can be used to scale up \first\ automatically.
\Cref{fig:orbit-vg} shows two examples of question-answer pairs generated automatically from Visual Genome (VG) \cite{krishna2017visual}, a popular dataset with high-quality structured connections between image regions, language descriptions, and WordNet \cite{miller1995wordnet} synsets. To investigate whether the annotations of VG can support the generation of large-scale instantiations of \benchmark, we devised a procedure that generates taxonomic questions by manipulating the WordNet synsets. This experiment exposed three limitations: lack of contextualization of classes (e.g., kitchen utensil instead of a bowl in \cref{fig:orbit-vg}-left), incorrect questions or answers (e.g., about uncountable objects like grass in \cref{fig:orbit-vg}-right), and classes in the WordNet hierarchy that are not overly specific (e.g., derby horse race) or generic (e.g., entity). Thus, while every image in VG enables many taxonomic questions to be created, it remains challenging to automatically distinguish high- from low-quality question-answer pairs. More fundamentally, creating question-answer pairs for other object properties (e.g., functional) and reasoning types (e.g., counterfactual) is hindered by the absence of annotation. While LLMs can be used to generate questions for these categories, we found that the resulting questions and answers are often incorrect or ambiguous (e.g., suggesting that buildings, doors, and windows can all provide shelter). In conclusion, we find that densely annotated image datasets remain insufficient for generating high-quality, large-scale versions of \benchmark. Yet, they could potentially be used to speed up human annotation by providing an initial set of samples or for distant supervision of VQA models.

\end{document}